\documentclass[journal]{IEEEtran}

\usepackage[font=footnotesize]{caption}
\usepackage{amsmath, bbm}
\usepackage[utf8]{inputenc} 
\usepackage[T1]{fontenc}    
\usepackage{hyperref}       
\usepackage{url}            
\usepackage{booktabs}       
\usepackage{amsfonts}       
\usepackage{nicefrac}       
\usepackage{microtype}      
\usepackage{lipsum}
\usepackage{fancyhdr}
\usepackage{gensymb}

\usepackage{float}
\usepackage{booktabs}
\usepackage{caption}
\usepackage{subcaption}
\usepackage{tabularx}
\usepackage{multirow}
\usepackage{siunitx}       
\usepackage{pgfplotstable} 
\usepackage{longtable}     
\usepackage{threeparttable}

\usepackage{graphicx}       
\graphicspath{{media/}}     

\usepackage{tikz}
\usetikzlibrary{shapes.geometric, arrows, positioning}

\tikzstyle{block} = [rectangle, draw, text centered, minimum height=3em]
\tikzstyle{arrow} = [thick,->]
\usepackage[table]{xcolor}   
\usepackage{arydshln}  

\usepackage[labelfont=bf]{caption} 
\definecolor{SecondBg}{RGB}{235,241,255}
\definecolor{BestBg}{RGB}{214,228,255}

\newcommand{\best}[1]{\cellcolor{BestBg}\textbf{#1}}
\newcommand{\second}[1]{\cellcolor{SecondBg}#1}

\newcommand{\tablesize}{\small}
\newcommand{\tightspacing}{\setlength{\tabcolsep}{5pt}}
\usepackage{bm}

\usepackage{ragged2e} 
\newcolumntype{Y}{>{\RaggedRight\arraybackslash}X}

\title{Scalable Context-Aware Graph Attention for Unsupervised Anomaly Detection in Large-Scale Mobile Networks}

\author{Sara Malacarne$^{1}$, Eirik Hoel-H\o iseth$^{1}$, Erlend Aune$^{3,5}$, David Zsolt Bir\'o$^2$, Massimiliano Ruocco$^{3,4}$
\thanks{This work was supported by the Norwegian Research Council projects ML4ITS (312062) and SFI NorwAI (309834).}
\vspace{.3cm}\\
1- Telenor Research and Innovation, Norway \\
2- Telenor Denmark OSS \& Tech Analytics, Denmark \\
3- Norwegian University of Science and Technology (NTNU), Norway \\
4- SINTEF Digital, Norway \\
5- Hance, Norway
}

\begin{document}
\maketitle

\thispagestyle{empty}

\begin{abstract}
Mobile network operators must monitor thousands of heterogeneous network elements across the
radio access network and the packet core, each exposing high-dimensional KPI time series. The scale
and cost of incident labelling make supervised approaches impractical, motivating unsupervised
anomaly detection robust to context shifts and nonstationarity.

We propose \textbf{C-MTAD-GAT} (\emph{Context-aware Multivariate Time-series Anomaly Detection with
Graph Attention}), an anomaly detection framework designed to operate as a single shared model across large
populations of network elements. The model combines temporal and feature-wise graph attention with
lightweight static and dynamic context conditioning and a dual-head decoder for reconstruction and
multi-step forecasting. It produces per-element, per-feature anomaly scores, converted to alerts via
fully unsupervised thresholds calibrated from validation residuals.

On the TELCO dataset released with DC-VAE \cite{garcia2023onemodel}, C-MTAD-GAT improves event-level affiliation and
pointwise F1 while generating fewer alarms than prior graph-attention and VAE-based baselines. We
then apply the same system to nation-scale radio access and evolved packet core control-plane counter
data from a mobile network operator, where it is deployed. Operator feedback indicates the alerts are
actionable and support daily monitoring, showing scalability across domains without relying on labelled incidents.

\end{abstract}


\section{Introduction}

Modern mobile networks are monitored through thousands of Key Performance Indicators (KPIs) collected from heterogeneous network elements (NEs) across multiple layers, including the Radio Access Network (RAN) and the Evolved Packet Core (EPC). Detecting abnormal behaviour early is operationally critical, but remains challenging because deviations can be subtle, distributed, and context-dependent.

Three constraints make this setting challenging. First, KPI streams are \emph{high-dimensional}, \emph{temporally dependent}, and \emph{interdependent}: a single fault may manifest as correlated deviations across multiple KPIs and NEs. Second, ground-truth labels are expensive to obtain and intrinsically hard to standardise, so anomaly detection (AD) must be largely unsupervised. Third, national networks involve thousands of NEs and hundreds of KPIs, making one-detector-per-NE approaches operationally infeasible; operators instead need \emph{centralised} models, i.e., single-model-per-domain, that scale across heterogeneous NEs while remaining compact and easy to retrain.
At this scale, AD is fundamentally a systems problem: beyond detection accuracy,
operators must address model centralisation, retraining cost, inference stability, and alert-volume
control under heterogeneous network configurations.

Most deep unsupervised AD methods detect anomalies via reconstruction or forecasting residuals, but operational deployments add further requirements: heterogeneity across NEs, metadata-driven context shifts (e.g., site configuration), and calibration without labels. Graph-attention-based models have proven effective at capturing inter-variable dependencies in
multivariate time series. However, prior work largely focuses on modelling accuracy and rarely
addresses how such models can be operated as unified, deployment-oriented systems that support
centralised training at scale, systematic use of contextual metadata, and fully unsupervised
calibration. Representative examples include MTAD-GAT~\cite{mtad-gat}.

In this work, we take a unified view across three settings: (i) the public TELCO benchmark with per-KPI labels; (ii) a large RAN dataset from a national 4G / LTE (Long-Term Evolution) network with thousands of sectors and rich context; and (iii) EPC control-plane counter data from virtualised core gateway units, where the system is currently deployed in the EPC. These settings share multivariate KPI time series but differ in labelling, dimensionality, and heterogeneity; we therefore target a \emph{single} architecture and a fully unsupervised calibration protocol that transfer across them.

We propose \textbf{C-MTAD-GAT}, a centralised context-aware graph-attention model that injects static and dynamic metadata via lightweight context conditioning. The model produces per-NE, per-feature anomaly scores, with thresholds estimated from validation errors only.

\noindent\textbf{Contributions.}
\begin{itemize}
    \item \textit{Centralised AD system:} We introduce C-MTAD-GAT, a context-aware
graph-attention-based system that enables a single shared anomaly detection model to operate across
thousands of heterogeneous telecom network elements.

    \item \textit{Calibration:} We present a domain-agnostic calibration protocol based solely on validation errors.
    \item \textit{Multi-domain validation:} We evaluate on TELCO, large-scale RAN, and EPC control-plane datasets and report deployment and human-in-the-loop considerations from a live EPC deployment.
    \item \textit{Scalability:} We analyse scaling with the training population size and characterise the stability of anomaly scores as the number of NEs increases toward operator scale.
\end{itemize}

\section{Background and State of the Art}
\label{sec:sota}

This section summarises deep unsupervised AD for multivariate time series, with emphasis on (i) residual-based scoring, (ii) dependency modelling across variables and time, and (iii) converting continuous scores into operational alerts without reliable labels.

\paragraph{Deep unsupervised multivariate time series AD}
Most deep unsupervised  time series AD approaches learn a model of normal dynamics and score anomalies through deviations, typically via reconstruction and/or forecasting residuals.
Classical deep baselines include sequence autoencoders and encoder--decoder models \cite{LSTM-ED}, variational approaches such as OmniAnomaly \cite{OmniAnomaly}, and GAN-based variants \cite{mad-gan}.
More recent work has increasingly relied on attention and transformer-style backbones to capture long-range temporal structure and complex cross-variable interactions, e.g., anomaly scoring via attention discrepancy in Anomaly Transformer \cite{xu2022anomalytransformer} and contrastive dual-attention designs such as DCdetector \cite{yang2023dcdetector}.
In addition, decomposition-guided designs have been proposed for online AD and forecasting (e.g., OneShotSTL in PVLDB) \cite{he2023oneshotstl}.
In practice, these backbones produce real-valued anomaly scores; downstream alert quality depends heavily on aggregation and thresholding choices, not only on architecture.

\paragraph{Modelling dependencies across variables and time}
In multivariate telemetry, anomalies often manifest as correlated deviations across subsets of KPIs.
A large class of methods therefore explicitly models dependencies, either through learned attention or graph structure.
Graph-based approaches can treat KPIs as nodes and learn inter-variable relations; MTAD-GAT is a representative example that couples temporal modelling with feature-wise graph attention for multivariate time series AD \cite{mtad-gat}.
Recent work further strengthens this direction with correlation-aware spatial--temporal graph learning \cite{zheng2023correlationaware}.
Transformer-based methods can capture dependencies implicitly through self-attention \cite{xu2022anomalytransformer}, often improving performance when interactions are non-local in time.

\paragraph{Score calibration and alerting without labels}
A core operational step is turning continuous scores into discrete alerts under limited or absent ground truth.
Common strategies include parametric thresholds under simple error models
(e.g., Gaussian/z-score rules based on mean and standard deviation),
robust statistics (e.g., median-based rules and median absolute deviation),
and tail modelling via Extreme Value Theory (EVT).
A widely used EVT baseline is Peaks-Over-Threshold (POT) and its streaming variants (SPOT/DSPOT), which fit a tail model to score exceedances above a high threshold \cite{siffer2017anomaly}.
More recently, the community has also emphasised that time series AD performance is sensitive to the end-to-end configuration (preprocessing, scoring, aggregation, thresholding), motivating automated selection and calibration pipelines such as AutoTSAD (PVLDB) \cite{autotsad2024}.
These works highlight that calibration is not an implementation detail but a first-class design choice for deployable AD systems.

\paragraph{Telecom-oriented studies and remaining gaps}
Telecom monitoring introduces operator-specific constraints, including high dimensionality, heterogeneous NEs, frequent regime shifts, and scarce or inconsistent incident labels. Prior work spans lightweight fault-management pipelines based on alarm correlation and rule-based reasoning in network operations \cite{li2010novel}, classical machine-learning ensembles \cite{oleiwi2022mlts}, transformer-based forecasting with adaptive thresholding \cite{zheng2023transks}, and multi-scale deep generative models such as dilated-convolutional VAEs \cite{garcia2023onemodel}.
However, fewer studies address the combined requirements typical of operator deployments: (i) \emph{centralised} training across large, heterogeneous NE populations, (ii) systematic use of operational side-information (context/exogenous covariates) to stabilise a single shared model, and (iii) alert calibration that is label-free yet produces actionable alarm volumes.
These gaps motivate our context-aware, centralised design and the evaluation across TELCO, RAN and EPC domains.

\section{Problem setup and notation}
\label{sec:problem}

We observe network elements (NEs) indexed by $n\in\{1,\dots,N\}$. For each NE $n$ we have a $k$-variate KPI time series
$X_n=\{\bm x^n_1,\ldots,\bm x^n_T\}$ with $\bm x^n_t\in\mathbb R^k$. Optionally, each NE has static metadata $\bm s^n$
(time-invariant) and time-aligned context $\bm z^n_t$ (Sec.~\ref{sec:preproc}).

We use inclusive slices: $\bm{x}^n_{a:b}:=(\bm{x}^n_a,\bm{x}^n_{a+1},\ldots,\bm{x}^n_b)$, so $|\bm{x}^n_{a:b}|=b-a+1$.
For window length $L$ and forecast horizon $H$, define the KPI window and aligned context at time $t$ by
\[
\bm X^n_{t-L+1:t}\in\mathbb R^{L\times k},
\;
\bm C^n_{t-L+1:t}:=\big(\bm s^n,\bm z^n_{t-L+1:t}\big)\in \mathcal S \times \mathbb R^{L\times d}.
\]

A centralised detector $g_\theta$ is trained on windows pooled across NEs and outputs a reconstruction $\tilde{\bm X}^n_{t-L+1:t}$
and an $H$-step forecast $\widehat{\bm X}^n_{t+1:t+H}$. 
From these we compute per-feature residuals and an anomaly score $\bm{e}^n_t \in \mathbb{R}^k$ given by the weighted combination of reconstruction/forecast errors.
We report both real-valued scores and binary decisions
$\bm y^n_t=\mathbbm 1\{\bm e^n_t>\bm\tau^n\}\in\{0,1\}^k$, where $\bm\tau^n$ is estimated in an unsupervised manner (Sec.~\ref{subsec:thresholding}) from validation residuals, per NE and feature. $L$ and $H$ are fixed within a domain (TELCO, RAN, EPC) and time splits are strictly time-ordered.

\section{Datasets}
\label{sec:data}

\subsection{TELCO Dataset -- Public Benchmark}
TELCO (TELeCOmmunication-networks) is a multivariate time series dataset from a live production
mobile network~\cite{garcia2023onemodel}. It provides $k{=}12$ KPIs (TS1–TS12) at 5-minute cadence over seven months (Jan~1–Jul~31, 2021), with per-series anomalous events manually labelled
by Network Operation Centre (NOC) experts. We follow the original time-ordered split into
train, validation and test sets. Anomalies are rare: on the test period there are $25\,143$
timestamps per series and $3\,001$ anomalous labels in total
($\approx 1\%$ positives), with per-KPI event counts between $1$ and
$35$.
\subsection{RAN Dataset -- Real data from a National MNO}
The dataset comes from a national mobile network operator (MNO) and covers the RAN, which connects user devices to the EPC via sectorised base stations.
In the operator network under study, there are approximately~$4\,000$ base stations, typically split into three $120^\circ$ sectors, yielding $\mathcal{O}(10^4)$ sectors.
We treat each sector as one monitored unit and aggregate KPIs at sector level, i.e., one multivariate KPI time series per sector.

Sectors are additionally grouped into coarser \emph{local areas}, operationally defined geographical clusters of nearby sectors, represented by the \texttt{local\_area} field.
Data are sampled hourly and include throughput, drop/setup failure rates, handover success, physical resource block (PRB) load, traffic volumes, and related counters.
Alongside dynamic real-valued KPIs, we use dynamic categorical context (e.g., hour of day, day of week/weekend, missingness flags) and static identifiers (e.g., sector and site IDs; optional vendor and band).

Experiments monitor around~$12\,000$ sectors using windows of $L{=}48$ input steps and $H{=}8$ forecast steps.
Representative variables are listed in Table~\ref{tab:ran_features}. Anomalies in the RAN are unlabelled; evaluation relies on expert validation (Sec.~\ref{sec:evaluation}).

\begin{table}[t]
\scriptsize
\tightspacing
\centering
\caption{Selected features used for AD in the RAN dataset. Acronyms: CS/PS = circuit-/packet-switched; PRB = physical resource block; RRC = radio resource control; UE = user equipment, IMSI = International Mobile Subscriber Identity.}
\begin{tabular}{|l|p{4.8cm}|}
\hline
\textbf{Type} & \textbf{Description} \\
\hline
\multicolumn{2}{|c|}{\textit{Selected dynamic real-valued variables}} \\
\hline
Real   & Duration the cell was available during the hour \\
Real   & Number of CS related drops \\
Real   & Call setup failure rate for CS traffic \\
Real   & Number of PS  related drops \\
Real   & Data setup failure rate \\
Real   & Downlink cell throughput \\
Real   & Uplink cell throughput \\
Real   & Downlink traffic volume (GB) \\
Real   & Uplink traffic volume (GB) \\
Real   & PRB load in downlink \\
Real   & Average number of RRC connected UEs \\
Real   & Maximum number of RRC connected UEs \\
Real   & Count of transmission-network-affected IMSIs \\
\hline
\multicolumn{2}{|c|}{\textit{Static real-valued variables}} \\
\hline
Real   & Sector latitude and longitude \\
Real   & Sector-level mean and standard deviation statistics per KPI \\
\hline
\multicolumn{2}{|c|}{\textit{Categorical variables}} \\
\hline
Static  & Encoded sector identifier \\
Dynamic & Day of week [0--6; 0=Mon] \\
Dynamic & Weekend flag (1=yes) \\
Dynamic & Hour of day [0--23] \\
Dynamic & Missing-value flag for PS  denominator \\
Dynamic & Missing-value flag for CS  denominator \\
\hline
\end{tabular}

\label{tab:ran_features}
\end{table}

\subsection{EPC Dataset -- Real data from a National MNO}
\label{sec:core-dataset}
Control-plane counters and KPIs from the EPC gateway units are collected at 5-minute cadence across two hosts of Packet Data Network Gateway (PGW).
Each record contains 205 dynamic real-valued counters covering session creation, bearer
modification (including VoLTE call sessions), failures, timeouts, and latencies,
plus 4 dynamic categorical time features (day of the week, weekend, hour of the day,
minutes of the hour).
Selected variables are shown in Table~\ref{tab:core_features}. Also this domain is unlabelled; evaluation relies on expert validation (Sec.~\ref{sec:evaluation}).

\begin{table}[t]
\scriptsize
\tightspacing
\centering
\caption{Selected features used for AD in the EPC dataset. Acronyms: PGW = packet data network gateway; PDP = packet data protocol; QCI = QoS class identifier.}

\label{tab:core_features}
\begin{tabular}{|l|p{4.8cm}|}
\hline
\textbf{Type} & \textbf{Description} \\
\hline
\multicolumn{2}{|c|}{\textit{Selected dynamic real-valued variables}} \\
\hline
Real   & PGW PDP-creation failure rate \\
Real   & Number of PGW PDP-creation rejects \\
Real   & Number of session-creation rejects \\
Real   & Number of bearer-modification failures \\
Real   & Number of session-deletion rejects \\
Real   & Number of idle-timeout closes \\
Real   & Number of error-indication messages \\
Real   & Mean disconnect latency \\
Real   & Bearer-setup failures (QCI 9) \\
\hline
\multicolumn{2}{|c|}{\textit{Categorical variables}} \\
\hline
Dynamic & Day of week [0--6; 0=Mon] \\
Dynamic & Weekend flag (1=yes) \\
Dynamic & Hour of day [0--23] \\
Dynamic & Minute bucket [0--59] \\
\hline
\end{tabular}

\end{table}

\section{Common preprocessing}
\label{sec:preproc}

Our input consists of both real-valued and categorical data. Incorporating both real and categorical features
allows us to effectively differentiate between network equipment while simultaneously
processing all data. We differentiate between four types of data per NE $n$ and time $t$: 
\begin{itemize}\itemsep 2pt
\item \textbf{Dynamic real} -- $\bm{x}^n_t\!\in\!\mathbb{R}^{k}$, time-variant KPIs, such as voice/data traffic
load, handovers, and error counters.
\item \textbf{Dynamic categorical} -- $\bm{z}^n_t$,  time-variant categorical flags such as day of the week,
hour, and missingness indicators.
\item \textbf{Static real} -- $\bm{r}^n$, time-invariant real-valued input such as fixed equipment
specifications, GPS coordinates or feature means and standard deviations (stds).
\item \textbf{Static categorical} -- $\bm{s}^n$, time-invariant categorical input such as sector, site, technology, frequency band.
\end{itemize}
We form a context vector
$\bm{C}^n_{t-L+1:t}=(\bm{s}^n,\bm{z}^n_{t-L+1:t})$ that will later condition the model
(Sec.~\ref{sec:method}).

\paragraph{Timeline alignment and splits}
We retain the native sampling cadence of each domain (TELCO/EPC: 5~min; RAN: 60~min).
To ensure sufficient temporal context, we apply training-time eligibility filters: NEs with
excessive missingness in the dynamic real KPIs (configurable; default $10\%$) or insufficient
history to support windowing are excluded from scaler/encoder fitting and from training. The
minimum history requirement is derived from $(L,H)$ by requiring at least two non-overlapping
windows in each non-empty split.

\paragraph{Missing values}
From the raw dynamic reals we compute a per-feature missingness mask
$\bm{m}^n_t\in\{0,1\}^k$ and append it as dynamic categorical indicators, so the model observes
both the filled signal and explicit missingness. We then fill remaining nulls with $0$
(for categoricals, the null is mapped to a dedicated token). No interpolation or
gap-dependent imputation is used.

\paragraph{Categorical encoding}
Static and dynamic categoricals are ordinal-encoded into $\mathbb{Z}_{\ge 0}$ using a single
encoder fit once per dataset.\footnote{For deployment this avoids refitting encoders when labels
are unavailable; for benchmarking one could fit on train only.} In the model, categorical
variables are mapped to learnable embeddings (Sec.~\ref{sec:method}).
\paragraph{Scaling}
Dynamic real KPIs are min--max scaled per NE and per feature using train-split statistics,
with a zero-range guard. Static real attributes are scaled globally across NEs using the
same scheme.

\paragraph{Windowing and time splits}
From each NE series we extract sliding windows $\bm{W}^n_t\in\mathbb{R}^{L\times k}$ with forecast horizon $H$ and stride $S$,
choosing valid start times so that $t{+}L{+}H{-}1$ lies within the series. Splits are strictly time-ordered (contiguous blocks) to
avoid temporal leakage and to mirror deployment. For RAN and EPC, the last $20\%$ of each NE timeline is reserved for test and
the remaining $80\%$ is split into train/val with an $80/20$ ratio.
For the TELCO dataset we use the predefined splits provided with the dataset. Windows are assigned to splits by time and we only keep
windows whose full input and forecast target lie entirely within the same split (i.e., no window crosses a split boundary). Each window
is paired with its aligned context $\bm{c}^n_t$ and an $H$-step forecast target.

\section{Method}
\label{sec:method}

\begin{figure*}[t]
    \centering
    \includegraphics[width=0.9\textwidth]{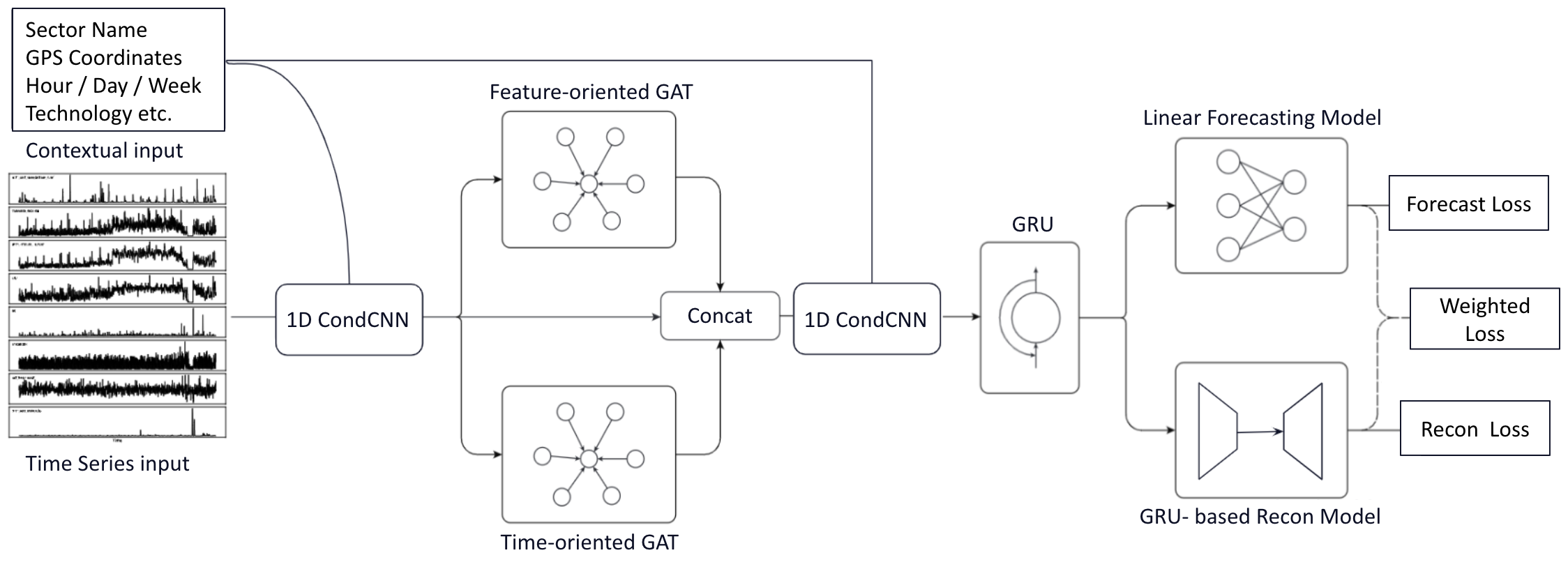}
    \caption{Architecture of C-MTAD-GAT for contextual and numerical input data.}
    \label{fig:c_mtadgat}
\end{figure*}

\subsection{Model architecture}
\label{subsec:model_architecture}

Given windowed KPI inputs $\bm{W}^n_t$ and aligned context $\bm{C}^n_t$
(Sec.~\ref{sec:problem}--\ref{sec:preproc}), C-MTAD-GAT extends MTAD-GAT
with (i) lightweight context conditioning, (ii) dynamic graph attention
(GATv2), and (iii) dual forecasting and reconstruction heads.
For each domain (TELCO, RAN, EPC) we train a single \emph{centralised}
model whose parameters are shared across all network elements (NEs), while
static and dynamic metadata modulate the shared backbone
(Fig.~\ref{fig:c_mtadgat}).
\paragraph{Context conditioning}
We embed categorical metadata (static and dynamic) with learnable
embeddings and project continuous static attributes (when available) to
the same latent space. These context representations condition the early
temporal feature extractor by modulating intermediate activations (a
FiLM \cite{perez2018film}/conditional-convolution style mechanism), allowing the shared model
to adapt to systematic differences across heterogeneous NEs (e.g., sector,
band, vendor, site) without training per-NE detectors.

\paragraph{Dependency modelling via graph attention}
From the context-conditioned temporal features, we form two
attention-enhanced views as in MTAD-GAT \cite{mtad-gat}:
(i) a \emph{feature-attention} module that treats KPI dimensions as nodes
and learns inter-variable dependencies (spatial graph attention), and
(ii) a \emph{temporal-attention} module that treats timesteps as nodes and
learns temporal dependencies (temporal graph attention).
We replace the original GAT layers with GATv2 \cite{brody2022how}, where
attention coefficients are a function of both source and target node
representations, improving expressiveness in directional and
context-sensitive interactions. Outputs from temporal features, feature
attention, and temporal attention are fused into a shared sequence
representation.

\paragraph{Sequence encoder and dual decoders}
A GRU encoder aggregates the fused representation over the window into a
latent state.
From this state we produce two outputs:
(i) a non-autoregressive multi-step forecasting head that predicts the next
$H$ steps for all KPIs in one forward pass, and
(ii) a deterministic GRU autoencoder head that reconstructs the input
window. Compared to MTAD-GAT's VAE head, the deterministic reconstruction
avoids sampling noise and KL-weight sensitivity, which we found to be
important for stable training and calibration in telecom monitoring.

\paragraph{Training objective and optimisation}
We train on sliding windows under the standard assumption that anomalies
are rare. The objective is a weighted sum of forecasting and
reconstruction losses (RMSE):
\[
\mathcal{L}=\mathcal{L}_{\text{forecast}}+\gamma\,\mathcal{L}_{\text{recon}},
\]
with $\gamma$ fixed per experiment.
Optimisation uses Adam with early stopping on validation loss. Domain-specific
settings (window length $L$, horizon $H$, stride $S$, batch size $B$, epochs
$E$) are in Tab.~\ref{tab:domain_hparams}, and Optuna-tuned architecture
hyperparameters are in Tab.~\ref{tab:best_hparams}.

\begin{table}[t]
  \centering
  \tablesize
  \tightspacing
  \caption{Per-domain data and training configuration for C-MTAD-GAT:
  window length $L$, forecast horizon $H$, stride $S$, batch size $B$, and
  maximum epochs $E$ used for TELCO, RAN and EPC.}
  \label{tab:domain_hparams}
  \scriptsize
  \begin{tabular}{lcccccc}
    \toprule
    \textbf{Domain} &
    \textbf{$L$} &
    \textbf{$H$} &
    \textbf{$S$} &
    \textbf{$B$} &
    \textbf{$E$} &
    \boldmath$\gamma$\unboldmath \\
    \midrule
    
    RAN   &24  &7  &17  &100  &30  &1  \\
    EPC   &101  &53 &89  &30  &200  &1  \\
    TELCO &577  &257  &31 &30  &150  &1  \\
    \bottomrule
  \end{tabular}
\end{table}

\begin{table}[t]
  \centering
  \tablesize
  \tightspacing
  \caption{Optimal Optuna-tuned architecture hyperparameters for C-MTAD-GAT on
  TELCO, RAN and EPC, including kernel sizes, GRU dimensions,
  layer widths and dropout rates.}
  \label{tab:best_hparams}
  \scriptsize
  \begin{tabular}{lccc}
\toprule
\textbf{Parameter} & \textbf{RAN} & \textbf{EPC} & \textbf{TELCO} \\
\midrule
Kernel size (1D conv)         & 4   & 4   & 18 \\
Use GATv2                     & True & True & True \\
GRU layers                    & 1   & 1   & 1 \\
GRU hidden dim                & 580 & 780 & 820 \\
Forecasting layers            & 3   & 1   & 4 \\
Forecast hidden dim           & 400 & 350 & 150 \\
Reconstruction layers         & 1   & 5   & 1 \\
Reconstruction hidden dim     & 400 & 800 & 150 \\
Dropout                       & 0.07 & 0.10 & 0.04 \\
Init.\ learning rate          & $2.487\times 10^{-4}$ & $2.488\times 10^{-4}$ & $1.728\times 10^{-4}$ \\
\bottomrule
\end{tabular}

\end{table}

\subsection{Unsupervised anomaly scoring and calibration}
\label{subsec:thresholding}

Our operational goal is fully label-free alerting: thresholds are derived
from validation residuals only (no tuning on incident labels), with
per-(NE, feature) flexibility and a simple, explainable rule.

For each NE $n$, feature $f$, and timestamp $t$, we compute a nonnegative
error by combining forecasting and reconstruction residuals (equal weight
in our experiments):
\[
e_{n,f,t}=\tfrac{1}{2}\bigl(e^{\text{for}}_{n,f,t}+e^{\text{rec}}_{n,f,t}\bigr).
\]
For each (NE, feature) pair we fit a simple parametric model to validation
errors and set a threshold as a high quantile. Concretely, we model
$\{e_{n,f,t}\}_{t\in\mathrm{VAL}}$ with an exponential distribution
$\mathrm{Exp}(\theta_{n,f})$, whose MLE scale is the sample mean
$\hat{\theta}_{n,f}$.
Given a target tail probability $p$ (fixed per domain), the threshold is
\[
\tau^n_f = -\hat{\theta}_{n,f}\ln(1-p).
\]
We flag anomalies when $e_{n,f,t}>\tau^n_f$. This yields a transparent
calibration mechanism with a single interpretable knob ($p$) that controls
alarm volume consistently across datasets.

\paragraph{Distributional sanity checks}
We validate the exponential-tail assumption by comparing it to Gamma fits
per (NE, feature) using likelihood-based criteria (LRT/AIC) and
out-of-sample log-likelihood on held-out data. While Gamma can provide
slightly heavier tails on RAN, the added complexity yields limited
practical benefit relative to the simplicity and operational stability of
the exponential calibration; therefore we use the exponential model in all
reported results.
For completeness, we also compare against EVT-style peaks-over-threshold
calibration on TELCO in Sec.~\ref{sec:results-telco}.

\section{Evaluation}
\label{sec:evaluation}

Evaluating AD in telecom networks is challenging: anomalies rarely have a clear start
or end point, may affect only a subset of KPIs with time lags across features, and labels are often
subjective or incomplete. In practice there is also a tension between \emph{early detection}
(operationally valuable but prone to false alarms) and \emph{full anomaly coverage}
(useful for post-mortem analysis but sometimes too late for mitigation). We therefore use
different strategies depending on the available supervision: event-level metrics on the TELCO
benchmark, and expert-guided evaluation on RAN and EPC.

\subsection{TELCO benchmark: metrics and setup}
\label{sec:telco-eval}

On the TELCO open benchmark we can use labelled anomalies and therefore adopt
standard quantitative metrics. We compare C-MTAD-GAT to three strong
baselines: MTAD-GAT, DC-VAE (dilated-convolutional VAE, the
strongest published TELCO baseline), and a $\beta$-MTAD-GAT variant
with down-weighted KL term ($\beta{=}0.2$). All models are trained on the same
TELCO splits and use the \emph{same} label-free per-feature thresholds on
validation errors, as described in Subsec.~\ref{subsec:thresholding}; TELCO labels
are reserved for evaluation only.

\paragraph{Pointwise timestamp-level metrics}
For each KPI $f$ we turn scores into binary timestamp labels and compute
precision (P), recall (R), and F1 at the \emph{timestamp} level. Because TELCO
is highly imbalanced, we also report a prevalence-matched \textbf{Random}
baseline: for each KPI, Random flags each timestamp as anomalous with
probability equal to that KPI’s empirical anomaly prevalence. This provides a
natural floor: any useful model should substantially outperform this baseline,
even if the resulting F1 scores look numerically small.

\paragraph{Event-wise affiliation}
Pointwise metrics are strict: missing a few timestamps in a long incident
heavily penalises recall. To better reflect incident-level performance, we also
report \emph{event-level affiliation} scores following \cite{huet2022local}.

For each feature $f$, consecutive positive timestamps are merged into
\emph{events}: ground-truth $G_f=\{g_k^f\}$ and predictions
$P_f=\{p_\ell^f\}$, where each event is a time interval. For any $g$ and $p$ we
define the intersection-over-union (IoU)
\[
\mathrm{IoU}(g,p)=\frac{|g\cap p|}{|g\cup p|},
\]
with $|\cdot|$ denoting duration. Each predicted event $p\in P_f$ is affiliated
to the ground-truth event $\arg\max_{g\in G_f}\mathrm{IoU}(g,p)$. We count a
true positive (TP) if $\max_{g\in G_f}\mathrm{IoU}(g,p)>0$, otherwise a false
positive (FP); any $g\in G_f$ not overlapped by any $p\in P_f$ is a false
negative (FN). Event-level P, R and F1 are then computed in the usual way.

Affiliation metrics are intentionally forgiving: on TELCO a trivial detector
that flags \emph{every} timestamp as anomalous still attains per-feature
affiliation F1 around $0.67$–$0.68$. Affiliation scores must therefore be
interpreted together with stricter pointwise metrics and the Random baseline to
avoid over-crediting overly active detectors.

\paragraph{Aggregation: Macro, Micro and Union}
We summarise TELCO performance in three complementary ways:

\begin{itemize}
  \item \textbf{Macro}: compute $(\mathrm{P}_f,\mathrm{R}_f,\mathrm{F1}_f)$
  separately for each KPI $f$ and report the unweighted average across
  features. This treats all KPIs equally, regardless of their frequency or
  number of events.

  \item \textbf{Micro}: first pool TP/FP/FN counts across all features, then
  compute precision, recall and F1 from these global totals. Micro scores
  therefore emphasise KPIs with many events.

  \item \textbf{Union}: take a logical OR across features at each timestamp to
  form single ground-truth and predicted streams, then apply the same
  pointwise or event-affiliation rules. Union scores capture “at least one KPI
  in trouble” behaviour, which is closer to how operators perceive incidents.
\end{itemize}

\subsection{RAN and EPC: expert-centred evaluation}
\label{sec:ran-core-eval}

In operational RAN and EPC environments we do not have dense, reliable anomaly labels. Many
statistically significant deviations are expected (e.g., traffic spikes during events) and not
business anomalies, while some subtle degradations are only evident in hindsight via trouble
tickets or configuration changes. For these domains we therefore combine quantitative proxies
with qualitative expert assessment.

\begin{figure*}[t]
  \centering

  \begin{subfigure}[t]{0.48\linewidth}
    \centering
    \includegraphics[width=\linewidth]{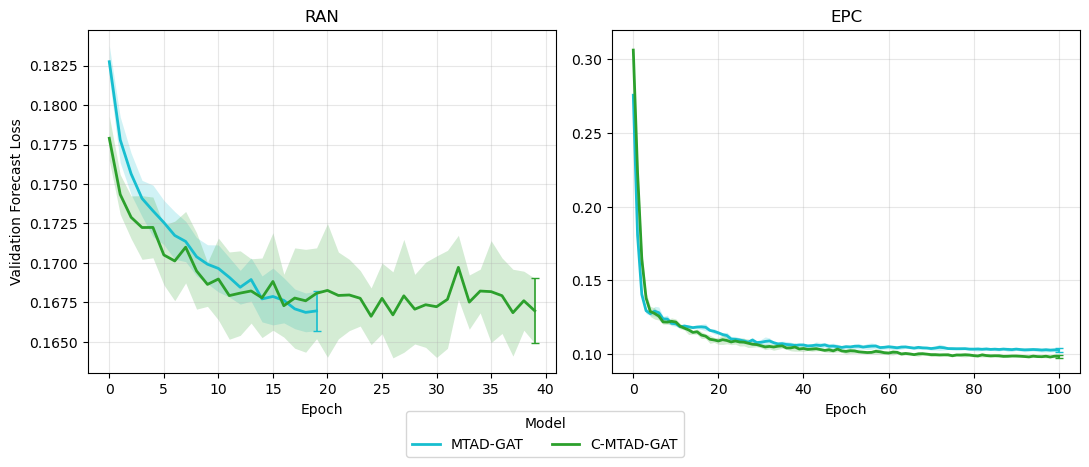}
    \caption{Validation \textbf{forecast} loss.}
    \label{fig:ran-core-forecast}
  \end{subfigure}
  \hfill
  \begin{subfigure}[t]{0.48\linewidth}
    \centering
    \includegraphics[width=\linewidth]{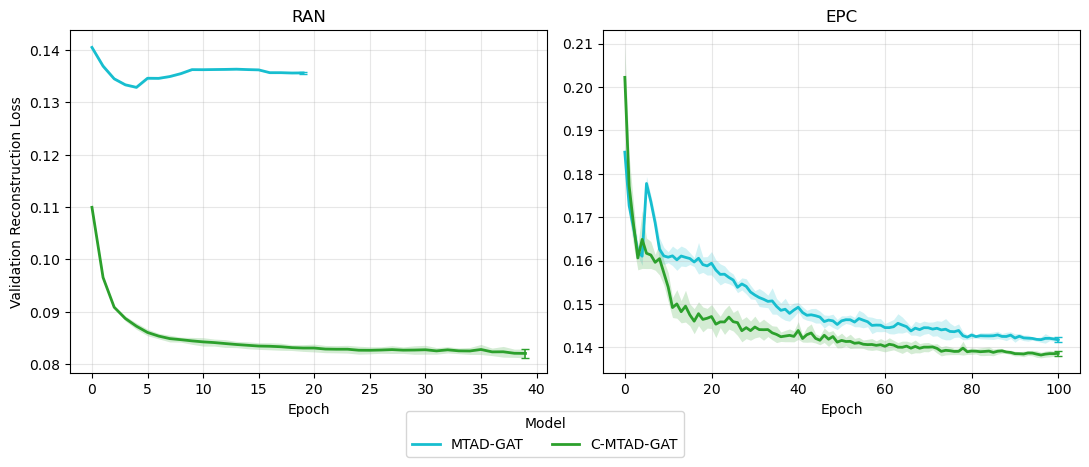}
    \caption{Validation \textbf{reconstruction} loss.}
    \label{fig:ran-core-recon}
  \end{subfigure}

  \scriptsize
  \caption{\textbf{RAN vs.\ EPC.} Left curves in each panel: RAN; right: EPC.
  Curves compare MTAD-GAT (baseline) and C-MTAD-GAT (ours). Mean across seeds with
  shaded $\pm$1 sd; last-epoch whisker shown when $n\!\ge\!2$. Lower is better.}
  \label{fig:ran-core-two-metrics}
\end{figure*}

\textit{Quantitatively}, we track validation forecast and reconstruction loss across seeds and
domains (Fig.~\ref{fig:ran-core-two-metrics}). Lower, more stable losses indicate that the model
better captures normal dynamics and typically correlate with smoother residuals and fewer spurious
alerts. \textit{Qualitatively}, we inspect time series, residuals, and alert timelines for a curated set of NEs and
periods, in collaboration with network operations experts. Anomalies are considered \emph{useful}
if they (i) coincide with periods of degraded performance, (ii) triggered or would plausibly
trigger investigations or corrective actions, or (iii) align with known incidents or configuration
changes. This expert-centred validation reflects how AD is actually consumed in
telecom operations and complements the benchmark-style TELCO metrics in
Sec.~\ref{sec:telco-eval}.

\section{Results and Discussion}
\label{sec:results}

We report results on three settings: the public TELCO benchmark, large-scale
RAN and EPC datasets from a a national MNO, and qualitative case studies
from live deployments. On TELCO, C-MTAD-GAT is compared against MTAD-GAT,
a $\beta$-MTAD-GAT variant, and the Telco-specific DC-VAE under identical
unsupervised calibration, using both event-level (affiliation) and
timestamp-level (pointwise) metrics. On RAN and EPC we analyse
validation losses across multiple seeds, ablate architectural choices
(GAT vs.\ GATv2, context blocks, AE vs.\ VAE head), and study how performance
scales with the number of network elements (NEs). Finally, we discuss
operational feedback and incident case studies illustrating how the model
behaves in production.

\subsection{TELCO benchmark: quantitative comparison}
\label{sec:results-telco}

We first evaluate C-MTAD-GAT on the public TELCO dataset, comparing it to
MTAD-GAT, a $\beta$-MTAD-GAT variant with down-weighted KL, and the
TELCO-specific DC-VAE baseline. All models use the same
train/validation/test splits and the same unsupervised calibration
protocol (per-feature thresholds derived from validation errors, without
tuning on TELCO labels; cf.\ Secs.~\ref{sec:method}--\ref{sec:evaluation}).
Architecturally, C-MTAD-GAT replaces MTAD-GAT's GAT by GATv2, injects
context via conditional convolutions, and uses a deterministic GRU
reconstruction head without KL regularisation.

Tables~\ref{tab:telco_results_affiliation} and
\ref{tab:telco_results_pointwise} report per-KPI F1 scores for event-level
\emph{affiliation} and timestamp-level \emph{pointwise} evaluation,
respectively. The corresponding Macro/Micro/Union aggregations, together
with the number of predicted events/timestamps, are shown in
Tables~\ref{tab:telco_results_affiliation_agg} and
\ref{tab:telco_results_pointwise_agg}. As discussed in Sec.~\ref{sec:telco-eval}, affiliation scores are more forgiving and can be inflated by trivially over-active detectors; they must therefore be interpreted together with stricter pointwise metrics.

\begin{itemize}
    \item \textbf{Affiliation event-level.}
Across Macro/Micro/Union aggregations in
Table~\ref{tab:telco_results_affiliation_agg}, C-MTAD-GAT attains the
highest affiliation F1, with DC-VAE consistently second. 
$\beta$-MTAD-GAT achieves higher precision but substantially lower recall,
while MTAD-GAT is even more conservative and misses many incidents.
C-MTAD-GAT offers a more balanced precision--recall profile: recall remains
close to DC-VAE's while precision is higher, yielding better F1 and fewer
alarms. On the union stream, our model raises far fewer incident windows
than DC-VAE while maintaining a higher F1, reducing the number of alarms
that operators must inspect in daily operations. Per-feature scores in
Table~\ref{tab:telco_results_affiliation} show that these gains are
distributed across KPIs rather than driven by a single outlier feature.
\item \textbf{Pointwise timestamp-level.}
Pointwise F1 scores in
Table~\ref{tab:telco_results_pointwise_agg} are lower in absolute value, as
expected under severe imbalance, but they act as a useful counterweight to
affiliation. C-MTAD-GAT achieves the best Macro and Union pointwise F1 and
matches DC-VAE on Micro F1, while firing fewer positives than both DC-VAE
and the prevalence-matched Random baseline. The Random row illustrates that
even moderate-looking F1 values correspond to clear gains over chance.
Per-feature pointwise scores in
Table~\ref{tab:telco_results_pointwise} confirm that improvements are
spread over multiple KPIs rather than isolated to a few “easy” ones.

\item \textbf{Backbone choice and robustness.}
The VAE-based MTAD-GAT variants require careful tuning of the KL weight and
additional regularisation tricks, and still underperform the simpler GRU-AE
head in C-MTAD-GAT under the same unsupervised calibration.
Standalone VAE variants (without graph attention; not shown) exhibited
similar sensitivity: performance degraded markedly away from hand-tuned
settings, which is problematic in monitoring pipelines that must be
retrained regularly under resource and reliability constraints.
In contrast, the deterministic GRU-AE backbone was robust: even
non–fully optimised hyperparameters yielded competitive performance, which
is crucial in large-scale operations.

\item \textbf{Thresholding robustness (POT vs.\ simple tails).}
For MTAD-GAT on TELCO we also compared our simple validation-based
thresholds against POT/SPOT-style extreme-value tails
(Table~\ref{tab:telco-pot-mtadgat}). In almost all KPIs, the simpler
per-feature thresholds derived from the fitted error distribution yielded
higher or comparable F1 while being easier to deploy and maintain. This
supports our choice of using a light-weight distributional model for errors
and rejecting POT/SPOT-style tails in the production setting.
\end{itemize}

\begin{table}[t]
  \centering
  \setlength{\tabcolsep}{2pt}
  \scriptsize
  \tightspacing
  
  \setlength{\tabcolsep}{3pt}

\begin{tabular}{l r c c c c c c c c}
\toprule
 & &
  \multicolumn{2}{c}{\textbf{C-MTAD-GAT}} &
  \multicolumn{2}{c}{\textbf{$\beta$-MTAD-GAT}} &
  \multicolumn{2}{c}{\textbf{MTAD-GAT}} &
  \multicolumn{2}{c}{\textbf{DC-VAE}} \\
\cmidrule(lr){3-4}\cmidrule(lr){5-6}\cmidrule(lr){7-8}\cmidrule(lr){9-10}
\textbf{Feature} & \textbf{GT} &
  Pred & F1 &
  Pred & F1 &
  Pred & F1 &
  Pred & F1 \\
\midrule
T1  & 20 &
  6  & \second{0.445} &
  1  & 0.095 &
  0  & 0.000 &
  19 & \best{0.538} \\
T2  & 35 &
  18 & \second{0.554} &
  3  & 0.142 &
  0  & 0.000 &
  59 & \best{0.599} \\
T3  &  7 &
  36 & 0.573 &
  28 & \second{0.584} &
  14 & \best{0.590} &
  33 & 0.579 \\
T4  &  8 &
  71 & \second{0.724} &
  62 & \best{0.744} &
  12 & 0.708 &
  79 & 0.678 \\
T5  &  8 &
  15 & \second{0.552} &
  8  & 0.521 &
  4  & 0.415 &
  12 & \best{0.621} \\
T6  &  7 &
  15 & 0.802 &
  10 & \second{0.814} &
  1  & 0.249 &
  31 & \best{0.844} \\
T7  & 10 &
  101 & \best{0.734} &
  73  & \second{0.709} &
  2   & 0.206 &
  178 & 0.683 \\
T8  &  8 &
  25 & \best{0.819} &
  19 & 0.773 &
  1  & 0.222 &
  34 & \second{0.773} \\
T9  & 17 &
  125 & \second{0.656} &
  128 & \best{0.693} &
  93  & 0.611 &
  125 & 0.655 \\
T10 & 19 &
  31 & \best{0.487} &
  29 & 0.426 &
  3  & 0.079 &
  34 & \second{0.468} \\
T11 &  3 &
  50 & \best{0.772} &
  40 & \second{0.765} &
  15 & 0.715 &
  60 & 0.636 \\
T12 &  1 &
  10 & \second{0.632} &
  28 & 0.614 &
  4  & \best{0.658} &
  49 & 0.553 \\
\bottomrule
\end{tabular}

  \caption{TELCO — \textbf{Affiliation (event-level)} F1 per feature. Leftmost GT = ground-truth events per feature. Each model block shows Pred (predicted events). Best is \textbf{bold}; second-best is \second{shaded}.}
  \label{tab:telco_results_affiliation}
\end{table}

\begin{table}[t]
\centering
\setlength{\tabcolsep}{2pt}
\scriptsize  
\renewcommand{\arraystretch}{1.1}
\begin{tabular}{lccc ccc ccc}
\toprule
& \multicolumn{3}{c}{\textbf{Macro}} 
& \multicolumn{3}{c}{\textbf{Micro}} 
& \multicolumn{3}{c}{\textbf{Union}} \\
\cmidrule(lr){2-4}\cmidrule(lr){5-7}\cmidrule(lr){8-10}
\textbf{Model} & P & R & F1 & P & R & F1 & P & R & F1 \\
\midrule
DC-VAE      & 0.644 & \best{0.682}   & \second{0.663}   &
              0.694 & \best{0.589}   & \second{0.637}   &
              0.561 & \best{0.798}   & \second{0.659}   \\
MTAD-GAT    & 0.589 & 0.339          & 0.430            &
              0.412 & 0.201          & 0.270            &
              \best{0.601} & 0.446  & 0.512            \\
$\beta$-MTAD-GAT
            & \best{0.716} & 0.591 & 0.647 &
              \best{0.793} & 0.420 & 0.549 &
              0.549 & 0.674 & 0.605 \\
\midrule
\textbf{C-MTAD-GAT}
            & \second{0.707} & \second{0.673} & \best{0.690} &
              \second{0.781} & \second{0.544} & \best{0.641} &
              \second{0.582} & \second{0.776} & \best{0.665} \\
\bottomrule
\end{tabular}

\captionsetup{font=footnotesize}
\caption{TELCO — \textbf{Aggregated Affiliation} scores. Best is \textbf{bold}; second-best is \second{shaded}. \textbf{Event Counts}. \emph{Micro}: Ground Truth (GT)=$143$, MTAD-GAT=$149$, $\beta$-MTAD-GAT=$429$, DC-VAE=$713$, \textbf{C-MTAD-GAT}=$503$. \emph{Union}: GT=$70$, MTAD-GAT=$133$, $\beta$-MTAD-GAT=$337$, DC-VAE=$567$, \textbf{C-MTAD-GAT}=$389$.}
\label{tab:telco_results_affiliation_agg}
\end{table}

\begin{table}[t]
  \centering
  \setlength{\tabcolsep}{2pt}
  \scriptsize
  \tightspacing
\setlength{\tabcolsep}{3pt}

\begin{tabular}{l r c c c c c c c c c}
\toprule
 & &
  \multicolumn{2}{c}{\textbf{C-MTAD-GAT}} &
  \multicolumn{2}{c}{\textbf{$\beta$-MTAD-GAT}} &
  \multicolumn{2}{c}{\textbf{MTAD-GAT}} &
  \multicolumn{2}{c}{\textbf{DC-VAE}} &
  $\boldsymbol{\pi_f}$ \\
\cmidrule(lr){3-4}\cmidrule(lr){5-6}\cmidrule(lr){7-8}\cmidrule(lr){9-10}
\textbf{Feature} & \textbf{GT} &
  Pred & F1 &
  Pred & F1 &
  Pred & F1 &
  Pred & F1 &
  \\
\midrule
T1  & 412 &
  7   & \second{0.029} &
  2   & 0.010 &
  0   & 0.000 &
  21  & \best{0.055} &
  0.016 \\
T2  & 501 &
  46  & \best{0.143} &
  4   & 0.016 &
  0   & 0.000 &
  107 & \second{0.115} &
  0.020 \\
T3  & 138 &
  71  & 0.019 &
  53  & \second{0.021} &
  25  & \best{0.025} &
  58  & 0.020 &
  0.005 \\
T4  & 179 &
  91  & \second{0.156} &
  80  & 0.154 &
  14  & 0.052 &
  99  & \best{0.165} &
  0.007 \\
T5  & 106 &
  43  & 0.268 &
  28  & \best{0.299} &
  12  & 0.186 &
  35  & \second{0.284} &
  0.004 \\
T6  & 107 &
  33  & \best{0.300} &
  23  & \second{0.262} &
  1   & 0.019 &
  60  & 0.251 &
  0.004 \\
T7  & 235 &
  129 & \second{0.115} &
  102 & \best{0.119} &
  4   & 0.025 &
  231 & 0.103 &
  0.009 \\
T8  & 237 &
  55  & \best{0.164} &
  40  & \second{0.159} &
  1   & 0.008 &
  70  & 0.156 &
  0.009 \\
T9  & 561 &
  453 & 0.030 &
  479 & \second{0.033} &
  341 & 0.024 &
  447 & \best{0.036} &
  0.022 \\
T10 & 489 &
  93  & \best{0.027} &
  90  & \second{0.024} &
  9   & 0.000 &
  91  & 0.010 &
  0.019 \\
T11 &  21 &
  210 & 0.043 &
  177 & 0.030 &
  51  & \best{0.083} &
  171 & \second{0.063} &
  0.001 \\
T12 &  15 &
  158 & \best{0.035} &
  131 & 0.000 &
  18  & 0.000 &
  92  & 0.000 &
  0.001 \\
\bottomrule
\end{tabular}

  \caption{TELCO — \textbf{Pointwise} F1 per feature. Leftmost GT = ground-truth events per feature. Each model block shows Pred (predicted timestamps). Best is \textbf{bold}; second-best is \second{shaded}.}
\label{tab:telco_results_pointwise}
\end{table}

\begin{table}[t]
\centering
\scriptsize
\setlength{\tabcolsep}{2pt}
\renewcommand{\arraystretch}{1.1}
\begin{tabular}{l|ccc|ccc|ccc}
\toprule
& \multicolumn{3}{c|}{\textbf{Macro}} & \multicolumn{3}{c|}{\textbf{Micro}} & \multicolumn{3}{c}{\textbf{Union}} \\
\cmidrule(lr){2-4}\cmidrule(lr){5-7}\cmidrule(lr){8-10}
\textbf{Model} & P & R & F1 & P & R & F1 & P & R & F1 \\
\midrule 
Random  & 0.010 & 0.010 & 0.010 & 0.015 & 0.015 & 0.015 & 0.047 & \second{0.113} & 0.067 \\ \hdashline 
DC-VAE              & 0.220 & \second{0.096} & \second{0.134} & \second{0.127} & \best{0.063} & \best{0.084} & 0.103 & 0.110 & \second{0.107} \\
MTAD-GAT    & \second{0.350} & 0.028 & 0.052 & 0.078 & 0.012 & 0.021 & \second{0.112} & 0.042 & 0.061 \\
$\beta$-MTAD-GAT     & \best{0.385} & 0.071 & 0.120 & 0.111 & 0.045 & 0.064 & 0.097 & 0.081 & 0.088 \\
\midrule
\textbf{C-MTAD-GAT}      & 0.319 & \best{0.107} & \best{0.160} & \best{0.133} & \second{0.062} & \best{0.084} & \best{0.117} & \best{0.115} & \best{0.116} \\
\bottomrule
\end{tabular}
\caption{TELCO — \textbf{Aggregated Pointwise} scores. Best is \textbf{bold}; second-best is \second{shaded}. 
\textbf{Timestamp Counts}.
\emph{Micro}: Ground Truth (GT)$=3001$, Random=3001, DC-VAE=1482, MTAD-GAT=476, $\beta$-MTAD-GAT=1209, \textbf{C-MTAD-GAT}=1389. 
\emph{Union}: GT$=1186$, Random=2850, DC-VAE=1268, MTAD-GAT=445, $\beta$-MTAD-GAT=991, \textbf{C-MTAD-GAT}=1163.}
\label{tab:telco_results_pointwise_agg}
\end{table}

\begin{table}[t]
\centering
\scriptsize
\tightspacing
\caption{Telco results: per-feature precision (P), recall (R), and F1 for MTAD-GAT comparing the Exponential Mean-Std thresholding vs POT thresholding. \textbf{Bold} marks the higher value per row/metric. }
\label{tab:telco-pot-mtadgat}
\begin{tabular}{lcccccc}
\toprule
& \multicolumn{3}{c}{POT-threshold} & \multicolumn{3}{c}{Exp-threshold} \\
\cmidrule(lr){2-4}\cmidrule(lr){5-7}
Feature & P & R & F1 & P & R & F1 \\
\midrule
TS1  & 0.684 & \textbf{0.180} & \textbf{0.285} & \textbf{1.000} & 0.050 & 0.095 \\
TS2  & 0.372 & \textbf{0.130} & \textbf{0.193} & \textbf{1.000} & 0.076 & 0.142 \\
TS3  & \textbf{0.649} & \textbf{0.756} & \textbf{0.698} & 0.627 & 0.647 & 0.637 \\
TS4  & 0.682 & 0.518 & 0.589 & \textbf{0.690} & \textbf{0.962} & \textbf{0.803} \\
TS5  & 0.486 & \textbf{0.547} & 0.514 & \textbf{0.759} & 0.410 & \textbf{0.533} \\
TS6  & 0.648 & 0.389 & 0.486 & \textbf{0.886} & \textbf{0.714} & \textbf{0.791} \\
TS7  & 0.553 & 0.383 & 0.452 & \textbf{0.685} & \textbf{0.766} & \textbf{0.723} \\
TS8  & 0.603 & \textbf{0.643} & 0.622 & \textbf{0.957} & 0.640 & \textbf{0.767} \\
TS9  & 0.402 & 0.396 & 0.399 & \textbf{0.654} & \textbf{0.723} & \textbf{0.687} \\
TS10 & 0.514 & 0.221 & 0.309 & \textbf{0.622} & \textbf{0.288} & \textbf{0.394} \\
TS11 & 0.386 & 0.743 & 0.508 & \textbf{0.651} & \textbf{0.908} & \textbf{0.759} \\
TS12 & 0.374 & 0.524 & 0.437 & \textbf{0.626} & \textbf{0.683} & \textbf{0.653} \\
\addlinespace
\textbf{Macro} & 0.529 & 0.452 & 0.488 & \textbf{0.763} & \textbf{0.572} & \textbf{0.654} \\
\textbf{Micro} & 0.515 & 0.333 & 0.405 & \textbf{0.817} & \textbf{0.412} & \textbf{0.548} \\
\textbf{Union} & \textbf{0.589} & 0.477 & 0.527 & 0.550 & \textbf{0.653} & \textbf{0.597} \\
\bottomrule
\end{tabular}

\end{table}

\subsection{RAN and EPC datasets: quantitative ablations}
\label{sec:results-operator}

We evaluate C-MTAD-GAT and ablations on nation-wide \emph{RAN} and \emph{EPC} datasets.
For each configuration and seed, we log validation forecasting and reconstruction losses at every epoch and
retain the \emph{minimum over epochs} of each metric; tables report mean $\pm$ standard deviation of these
per-seed minima across seeds.

We treat \textbf{C-MTAD-GAT (AE, ctx)} as the fixed \emph{operational baseline} (the deployed model),
where \emph{AE} denotes a deterministic GRU autoencoder reconstruction head (no KL term) and
\emph{ctx} denotes full context injection into both temporal convolution blocks.
Row names in Tables~\ref{tab:core-min-only}--\ref{tab:ran-min-only} vary three factors:
(i) \textbf{backbone/attention} (\emph{MTAD-GAT}: GAT; \emph{C-MTAD-GAT}: GATv2),
(ii) \textbf{reconstruction head} (\emph{AE} vs \emph{VAE} with KL regularisation),
and (iii) \textbf{context injection}.
Legend: \emph{no ctx} disables context; \emph{ctx@block1/2} injects context only into the first/second temporal
convolution block; and \emph{ctx} injects context into both blocks.
For EPC, static metadata are not available, hence \emph{ctx} always refers to dynamic time-related categoricals
(e.g., hour/weekday and missingness flags), and variants that differ only in static-context wiring become
effectively identical.

\begin{figure}[!t]
  \centering

  \begin{subfigure}[t]{0.48\textwidth}
    \centering
    \includegraphics[width=\linewidth]{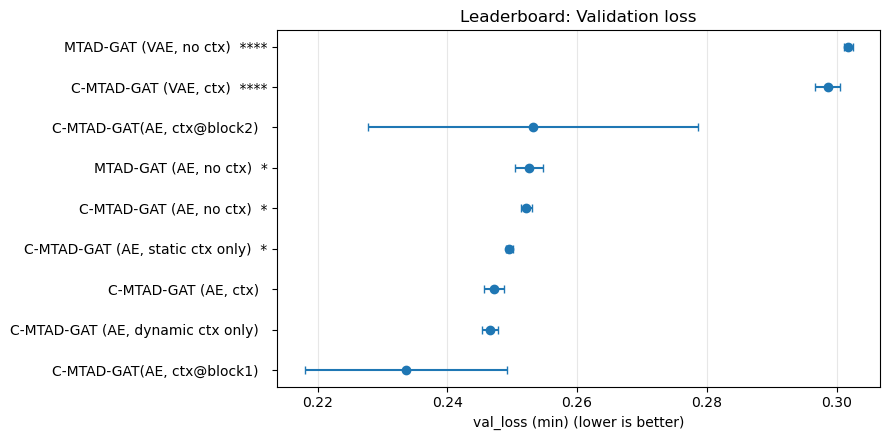}
    \caption{Leaderboard — RAN. Mean $\pm$ 95\% CI of per-seed values for the chosen metric (lower is better). Stars mark paired $t$-test significance vs the best. The best configuration appears at the top.}
    \label{fig:leaderboard-ran}
  \end{subfigure}\hfill
  \begin{subfigure}[t]{0.48\textwidth}
    \centering
    \includegraphics[width=\linewidth]{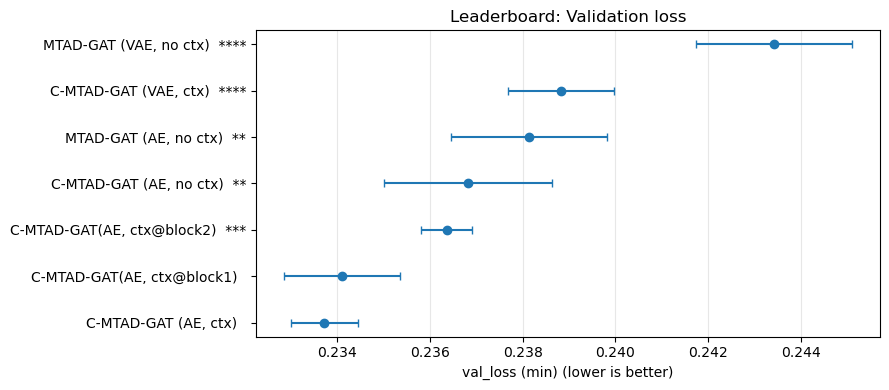}%
    \caption{Leaderboard — EPC. Same convention as (a): mean $\pm$ 95\% CI across seeds; stars indicate paired $t$-test $p$-values vs the best.}
    \label{fig:leaderboard-core}
  \end{subfigure}

  \medskip

  \begin{subfigure}[t]{0.48\textwidth}
    \centering
    \includegraphics[width=\linewidth]{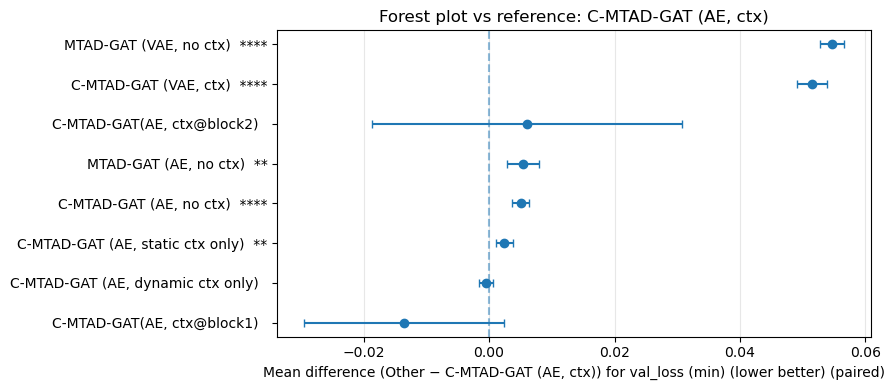}
    \caption{Forest — RAN. Each point is the paired mean difference (Other $-$ Best) across shared seeds, with 95\% CI; the vertical line at 0 indicates parity. Stars encode paired $t$-test $p$-values.}
    \label{fig:forest-ran}
  \end{subfigure}\hfill
  \begin{subfigure}[t]{0.48\textwidth}
    \centering
    \includegraphics[width=\linewidth]{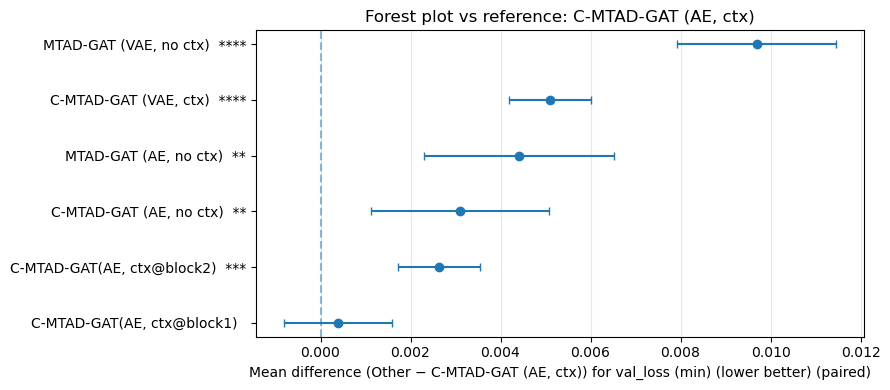}
    \caption{Forest — EPC. Mean differences relative to the best configuration, 95\% CI, and paired $t$-test significance as in (c).}
    \label{fig:forest-core}
  \end{subfigure}
\medskip
  \caption{Seed-robust ablation comparisons on \emph{RAN} and \emph{EPC} datasets. 
  Leaderboards ((a),(b)) summarize absolute performance (mean across seeds with 95\% CI; lower is better). 
  Forest plots ((c),(d)) show differences to the baseline model (Other $-$ baseline) with 95\% CI; values $>0$ mean the other model is worse. 
  Unless otherwise stated, all panels use the same metric and statistic (e.g., \texttt{val\_loss} with \texttt{min} across epochs).}
  \label{fig:leaderboards-forests}
\end{figure}

\begin{itemize}
    \item \textbf{Absolute vs.\ baseline performance.}
Tables~\ref{tab:core-min-only} and \ref{tab:ran-min-only} report absolute validation minima (lower is better).
To quantify robustness across seeds, Tables~\ref{tab:core-vs-best-valloss-min} and \ref{tab:ran_vs_fixed_val_loss_min}
report differences to the fixed baseline on \texttt{val\_loss|min}:
$\Delta = \text{Other} - \text{Baseline}$ (absolute and percent),
paired two-sided $t$-test $p$-values adjusted for multiple comparisons using the
Benjamini--Hochberg false discovery rate (BH--FDR; reported as $p_{\mathrm{adj}}$),
and Cohen's $d$ computed on paired differences (negative values favour the baseline).
Figure~\ref{fig:leaderboards-forests} visualises the same baseline-referenced comparisons.

\item \textbf{EPC.} 
On EPC, moving from MTAD-GAT (GAT) to C--MTAD-GAT (GATv2) with an AE head yields a modest but consistent improvement in
validation loss across seeds (Tables~\ref{tab:core-min-only} and \ref{tab:core-vs-best-valloss-min}).
VAE-based variants are systematically worse than AE variants, with significant degradations and larger effect sizes.
Because EPC lacks static metadata, context mechanisms are limited to dynamic time-related categoricals; accordingly, several C-MTAD-GAT AE variants are statistically indistinguishable from each other, with performance differences within numerical noise.
\item \textbf{RAN.}
On RAN, context-aware C-MTAD-GAT AE variants cluster near the baseline, while VAE variants degrade sharply with large
and significant gaps (Tables~\ref{tab:ran-min-only} and \ref{tab:ran_vs_fixed_val_loss_min}).
The best mean \texttt{val\_loss|min} is sometimes achieved by \emph{ctx@block1}, but this variant exhibits higher
across-seed variance (wider confidence intervals in Fig.~\ref{fig:leaderboards-forests}), suggesting sensitivity to
initialisation on heterogeneous RAN dynamics. In contrast, the full-context AE baseline is slightly more conservative
in mean loss but more stable across seeds.

\end{itemize}

\textbf{Operational takeaways.}
Across both operator datasets, (i) replacing GAT by GATv2 is never harmful and is often mildly beneficial,
(ii) context injection is most valuable on heterogeneous RAN data (dynamic context accounts for most of the gain),
and (iii) deterministic AE heads are substantially more stable than VAE variants under repeated retraining.
These results support the deployed design choice: C-MTAD-GAT with GATv2, context injection, and a GRU-AE
reconstruction head.

\begin{table}[t]
  \centering
  \scriptsize
  \tightspacing
  \caption{EPC ablation on validation forecasting and reconstruction loss
(mean $\pm$ SD of per-seed minima; lower is better). Best is \textbf{bold}, second-best is \second{shaded}.}

  \label{tab:core-min-only}










 \begin{tabular}{lcc}
  \toprule
  Configuration & val\_forecast\_loss & val\_recon\_loss \\
  \midrule
  MTAD-GAT (VAE, no ctx)            & 0.1021 $\pm$ 0.0015 & 0.1410 $\pm$ 0.0007 \\
  MTAD-GAT (AE, no ctx)             & 0.1001 $\pm$ 0.0016 & 0.1378 $\pm$ 0.0007 \\
  C-MTAD-GAT (AE, no ctx)        & 0.0993 $\pm$ 0.0018 & 0.1374 $\pm$ 0.0007 \\
  C-MTAD-GAT (AE, ctx@block1)          & \second{0.0971 $\pm$ 0.0010} & \second{0.1368 $\pm$ 0.0009} \\
  C-MTAD-GAT (AE, ctx@block2)         & 0.0981 $\pm$ 0.0005 & 0.1380 $\pm$ 0.0005 \\
  C-MTAD-GAT (VAE, ctx) & 0.0984 $\pm$ 0.0011 & 0.1401 $\pm$ 0.0005 \\
  C-MTAD-GAT (AE, ctx)  & \best{0.0968 $\pm$ 0.0004}   & \best{0.1367 $\pm$ 0.0008} \\
  
  \bottomrule
  \end{tabular}
\end{table}

\begin{table}[t]
  \centering
  \scriptsize
  \tightspacing
  \caption{RAN ablation on validation forecasting and reconstruction loss
(mean $\pm$ SD of per-seed minima; lower is better). Best is \textbf{bold}, second-best is \second{shaded}.}

  \label{tab:ran-min-only}










\begin{tabular}{lcc}
  \toprule
  Configuration & val\_forecast\_loss & val\_recon\_loss \\
  \midrule
  MTAD-GAT (VAE, no ctx)           & 0.1661 $\pm$ 0.0010 & 0.1329 $\pm$ 0.0002 \\
  MTAD-GAT (AE, no ctx)            & 0.1671 $\pm$ 0.0021 & 0.0846 $\pm$ 0.0014 \\
  C-MTAD-GAT (AE, no ctx)       & 0.1669 $\pm$ 0.0009 & 0.0844 $\pm$ 0.0006 \\
  C-MTAD-GAT (AE, ctx@block1)         & \best{0.1466 $\pm$ 0.0255} & 0.0865 $\pm$ 0.0055 \\
  C-MTAD-GAT (AE, ctx@block2)          & \second{0.1626 $\pm$ 0.0246} & 0.0896 $\pm$ 0.0124 \\
  C-MTAD-GAT (AE, dyn ctx only)           & 0.1636 $\pm$ 0.0015 & \best{0.0819 $\pm$ 0.0005} \\
  C-MTAD-GAT (AE, static ctx only)        & 0.1661 $\pm$ 0.0010 & 0.0826 $\pm$ 0.0004 \\
  C-MTAD-GAT (VAE, ctx)     & 0.1629 $\pm$ 0.0030 & 0.1305 $\pm$ 0.0001 \\
  C-MTAD-GAT (AE, ctx)      & 0.1640 $\pm$ 0.0018 & \second{0.0820 $\pm$ 0.0008} \\
  
  \bottomrule
  \end{tabular}
\end{table}
\begin{table}[t]
  \centering
  \scriptsize
  \tightspacing
  \caption{\textbf{EPC} ablations reported \emph{relative to a fixed baseline} 
C-MTAD-GAT (AE, ctx).
Each row reports the mean gap to the baseline (Other$-$baseline) in absolute units ($\Delta$)
and percent, computed on per-seed minima of \texttt{val\_loss|min} (lower is better).
$p_\mathrm{adj}$ are BH–FDR–adjusted two-sided paired $t$-test $p$-values;
\emph{Sig} encodes significance ($^{*}p{<}0.05$, $^{**}p{<}10^{-2}$,
$^{***}p{<}10^{-3}$, $^{****}p{<}10^{-4}$).
Cohen’s $d$ is computed on the paired differences (Other$-$baseline; positive favours the baseline).
}

  \label{tab:core-vs-best-valloss-min}
  \begin{tabular}{p{3.6cm}rrrcr}
  \toprule
  Configuration & $\Delta$ & $\Delta$ (\%) & $p_\mathrm{adj}$ & Sig & $d$ \\
  \midrule
  MTAD-GAT (VAE, no ctx)              & 0.0097 & 4.15 & $<10^{-4}$ & **** & 4.22 \\
  C-MTAD-GAT (VAE, ctx)               & 0.0051 & 2.18 & $<10^{-4}$ & **** & 4.35 \\
  MTAD-GAT (AE, no ctx)               & 0.0044 & 1.88 & $0.0013$   & **   & 1.60 \\
  C-MTAD-GAT (AE, no ctx)      & 0.0031 & 1.33 & $0.0068$   & **   & 1.20 \\
  C-MTAD-GAT (AE, ctx@block2)     & 0.0027 & 1.16 & $0.00015$  & ***  & 2.24 \\
  C-MTAD-GAT (AE, ctx@block1)     & 0.0004 & 0.17 & $0.49$     & —    & 0.24 \\
  \bottomrule
\end{tabular}

\end{table}


\begin{table}[t]
  \centering
  \tightspacing
  \scriptsize
  \caption{RAN ablations reported \emph{relative to a fixed baseline} 
C-MTAD-GAT (AE, ctx).
Each row reports the mean gap to the baseline (Other$-$baseline) in absolute units ($\Delta$)
and percent, computed on per-seed minima of \texttt{val\_loss|min} (lower is better).
$p_\mathrm{adj}$ are BH–FDR–adjusted two-sided paired $t$-test $p$-values;
\emph{Sig} encodes significance ($^{*}p{<}0.05$, $^{**}p{<}10^{-2}$,
$^{***}p{<}10^{-3}$, $^{****}p{<}10^{-4}$).
Cohen’s $d$ is computed on the paired differences (Other$-$baseline; positive favours the baseline).
}

\label{tab:ran_vs_fixed_val_loss_min}
\begin{tabular}{p{3.6cm}rrrcr}
  \toprule
  Config. & $\Delta$ & $\Delta$ (\%) & $p_\mathrm{adj}$ & Sig & $d$ \\
  \midrule
  MTAD-GAT (VAE, no ctx)              &  0.0546 & 22.1 & $<10^{-4}$ & **** & 22.7 \\
  C-MTAD-GAT (VAE,  ctx)          &  0.0515 & 20.8 & $<10^{-4}$ & **** & 16.8 \\
    C-MTAD-GAT (AE, ctx@block2)         &  0.0060 &  2.45 & $0.59$ & —    &  0.19 \\
  MTAD-GAT (AE, no ctx)               &  0.0054 &  2.20 & $0.0025$ & **   &  1.65 \\
  C-MTAD-GAT (AE, no ctx)      &  0.0050 &  2.03 & $<10^{-4}$        & **** &  2.88 \\
  C-MTAD-GAT (AE, static ctx only)    &  0.0024 &  0.97 & $0.0069$ & **   &  1.34 \\
  C-MTAD-GAT (AE, dyn ctx only) & -0.0005 & -0.23 & $0.35$ & —    & -0.39 \\
  C-MTAD-GAT (AE, ctx@block1)         & -0.0136 & -5.50 & $0.13$ & —    & -0.66 \\
  \bottomrule
\end{tabular}

\end{table}

\subsection{Scalability and stability vs.\ model granularity}
\label{sec:scalability-units}

A practical deployment question is whether to run one centralised model over
many NEs or many small per-NE models. 
To probe this trade-off on RAN, we
selected six representative sectors, including empirically challenging ones
where models show lower validation performance, and compared:
\begin{itemize}
  \item separate models trained on each of the six sectors individually
        (family \texttt{model\_1});
  \item a model trained jointly on the same six sectors (\texttt{model\_6});
  \item centralised models trained on increasingly large sets of sectors
        (50, 100, 500, 1000, 5000, 7000, and all sectors), always
        evaluating on the original six.
\end{itemize}

For each configuration and seed we record, for every sector, the minima over
epochs of the validation forecast and reconstruction losses as in
Sec.~\ref{sec:results-operator}. Across the six sectors, these validation
losses remain within a few percent when moving from per-sector to fully
centralised models, with no systematic degradation as more sectors are added.
Seed-to-seed variability is non-negligible: some large-model seeds still
outperform some small-model seeds and vice versa, but there is no evidence of
a collapse in performance as additional sectors are absorbed into a single
model.

\paragraph{Stability of anomaly patterns}
Validation loss alone does not tell us whether larger models raise similar
alarms. To quantify stability of the detected anomaly patterns, we consider,
for each sector and seed, the Jaccard overlap between the anomaly rows flagged
by the per-sector model (\texttt{model\_1}) and those flagged by a larger
family (e.g., \texttt{model\_50}). For a given sector and seed, let
$A_\text{baseline}$ and $A_\text{family}$ be the sets of anomaly keys; we
define
\[
  J = \frac{|A_\text{baseline} \cap A_\text{family}|}
           {|A_\text{baseline} \cup A_\text{family}|}.
\]
We then average $J$ across seeds and sectors, and report 95\% confidence
intervals over the same population.

\begin{figure}[t]
  \centering
  \begin{subfigure}{\linewidth}
    \centering
    \includegraphics[width=\linewidth]{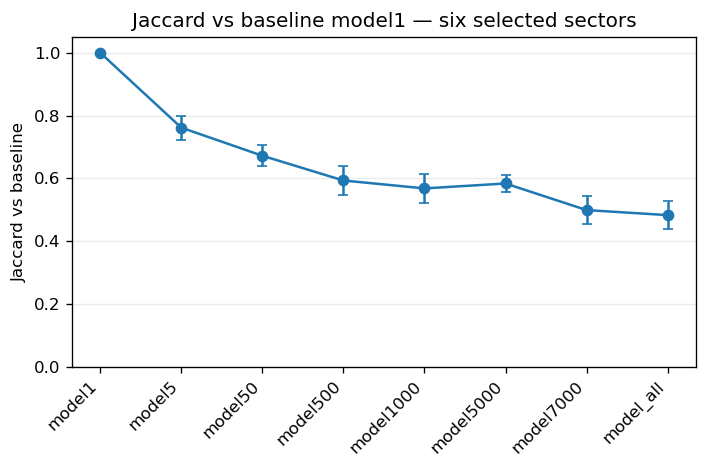}
    \caption{Mean Jaccard vs.\ model family on the six selected sectors.}
  \end{subfigure}

  \vspace{0.6em}

  \begin{subfigure}{\linewidth}
    \centering
    \includegraphics[width=\linewidth]{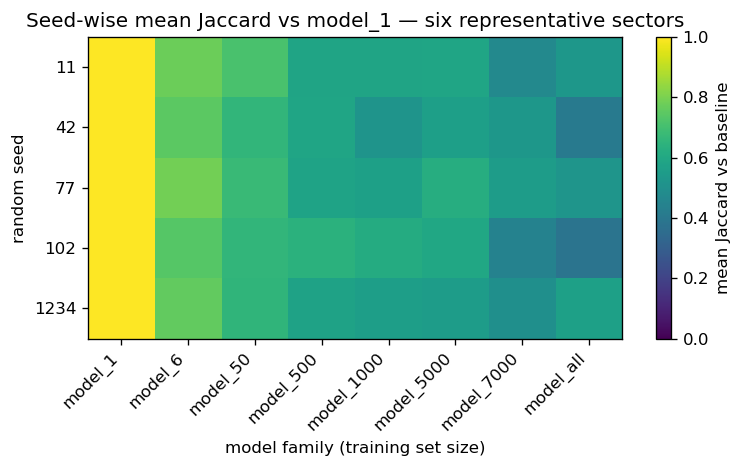}
    \caption{Seed–family heatmap: Jaccard between each centralised family and its per-sector baseline, across seeds.}
  \end{subfigure}

  \caption{Scalability and stability of anomaly patterns on six RAN sectors.
  (a) Jaccard overlap between anomaly rows from the per-sector model
  (\texttt{model\_1}) and larger centralised models, averaged over seeds;
  error bars denote 95\% confidence intervals across sectors. (b) For each
  family (columns) and seed (rows), Jaccard overlap between the centralised
  model and the corresponding per-sector model, aggregated over the same six
  sectors; values around $1$ indicate almost identical anomaly sets, values
  around $0.5$ indicate moderate drift.}
  \label{fig:jaccard-scalability}
\end{figure}

Figure~\ref{fig:jaccard-scalability}a shows how this Jaccard overlap evolves
as the training set grows, restricted to the six selected sectors. As expected,
the overlap decreases as we move from per-sector models towards large
centralised models trained on thousands of sectors: families trained on
50--500 sectors still agree with \texttt{model\_1} on roughly
$60$--$70\%$ of anomaly rows, whereas the all-sectors model keeps around
$45$--$50\%$ overlap. On our six representative sectors, growing the training
set from one sector to all sectors reduces the Jaccard overlap with the
per-sector model from $1$ to roughly $0.45$–$0.5$.

To contextualise this drift, we also examine the intrinsic seed-to-seed
variability of a fixed centralised model family (e.g.\ \texttt{model\_all}).
Figure~\ref{fig:jaccard-scalability}b shows, for each seed and family, the
Jaccard overlap between that centralised model and its corresponding
per-sector baseline, aggregated over the six sectors. For the fully
centralised family, these overlaps range from approximately $0.25$ to $0.75$
(mean $\approx 0.5$), which is of the same order as the degradation observed
when moving from single-sector to all-sectors training. This suggests that
the drift introduced by centralisation is comparable to the intrinsic
seed-to-seed variability of the model, rather than an additional, much
larger source of instability.

Taken together, the loss-based and Jaccard-based analyses indicate that, for
large RAN deployments, a single centralised C-MTAD-GAT model is viable in
practice and dramatically simpler to operate than thousands of small models.
Centralisation comes with a modest change in the anomaly sets, but without
catastrophic drift: even the all-sectors model retains a substantial fraction
of the per-sector alarms. In settings where specific outlier sectors are
business-critical, one can layer targeted fine-tuning or per-NE
specialisation on top of the centralised backbone.

\subsection{Operational Case Studies}
\label{sec:experiments}

In this section we move from aggregate metrics to concrete operational behaviour. We illustrate how the deployed C-MTAD-GAT behaves in a live EPC and RAN settings and how its outputs are used by network operations. Examples are drawn from the EPC and RAN datasets, described in Sec.~\ref{sec:core-dataset}. For business confidentiality, EPC host identifiers are anonymised (e.g., \texttt{host-10}, \texttt{host-50}), and charts show relative rather than absolute counter values.


\begin{table}[t]
\centering
\scriptsize
\setlength{\tabcolsep}{2pt} 
\caption{Illustrative RAN and EPC incidents surfaced by C-MTAD-GAT.}
\label{tab:incidents_ran_core}
\begin{tabularx}{\columnwidth}{p{1.0cm} p{2.2cm} Y Y}
\toprule
\textbf{Dom.} & \textbf{Counter} & \textbf{Root-cause summary} & \textbf{Operator action} \\
\midrule
RAN &
Transmission / backhaul stability KPIs &
External fault in the transmission provider’s network. &
Resolve transmission issue and monitor the affected area. \\
\midrule
EPC &
Update Bearer MME (Mobility Management Entity) Reject &
Inbound roamers; failed QCI/ARP modification on new MME. &
Adjust MME configuration (static QCI/ARP for roamers). \\
\midrule
EPC &
Maximal Durations Exceeded &
Stricter maximal call duration on new MME platform. &
Align VoLTE call-duration limits with previous platform. \\
\midrule
EPC & Create Session Reject (PGW not responding) &
Signalling storm after public-warning / cell-broadcast event. &
Tune memory allocation; monitor behaviour during/after event. \\
\midrule
EPC & Suspend Notification Reject &
Planned change to reduce RADIUS / MVNO signalling load. &
Acknowledge new baseline; no remediation needed. \\
\bottomrule
\end{tabularx}
\end{table}

\subsubsection{\textbf{Feedback from Network Operations}}

Feedback from operations engineers emphasized three recurring points:

\begin{itemize}
  \item \textit{Actionability over pure accuracy.}
  What mattered most was whether an alert led to a useful insight or
  operational decision. Precision in the 60--70\% range was considered
  acceptable if the remaining false positives were easy to inspect and
  helped maintain situational awareness.
  \item \textit{Multi-metric drift rather than single thresholds.}
  Engineers valued the ability to highlight subtle, multi-counter
  deviations that would not trigger traditional single-KPI thresholds,
  especially in the EPC domain where behaviour is distributed across
  layers and functions. This was seen as a scalable way to monitor complex
  platforms without an explosion of hand-crafted rules.
  \item \textit{Generalisation across heterogeneous infrastructure.}
  Categorical embeddings and context injection allowed a single model
  to operate across different hosts and configurations without per-host
  fine-tuning. This was seen as a prerequisite for practical deployment
  in large, evolving networks.
\end{itemize}

\subsubsection{\textbf{Operational Workflow and Prioritisation}}
\label{sec:operational-workflow}

To support day-to-day use, we introduced an \emph{aggregated anomaly priority}
view that combines anomaly scores over time, network elements (NEs), and counters.
The goal is not to replace per-counter plots but to provide operators with a
high-level, sortable overview of \emph{when} the system deviates and \emph{how
severe} the deviation is.

Concretely, for each counter we convert the normalised reconstruction/forecast
error into five discrete priority levels by binning the upper tail of that
counter's validation error distribution (level~5 = most extreme). We then
aggregate these priorities across NEs and counters at each time step to obtain
an \emph{aggregated priority} score (sum of per-counter priority levels), which
emphasises periods with widespread and/or extreme deviations.

In addition, we report the \emph{aggregated number of anomalies}, defined as the
count of (NE, counter) pairs whose error exceeds the anomaly threshold at each
time step (i.e., the number of active alarms, irrespective of priority).

\begin{figure}[t]
  \centering

  \begin{subfigure}{0.9\linewidth}
    \centering

\includegraphics[width=\linewidth]{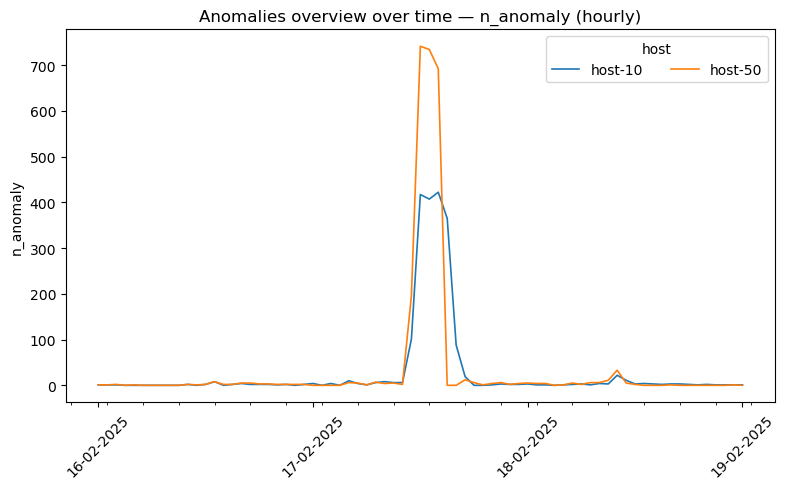}

    \caption{EPC (each line is a host).}
    \label{subfig:epc_agg}
  \end{subfigure}

  \vspace{0.6em}

  \begin{subfigure}{0.9\linewidth}
    \centering
    \includegraphics[width=\linewidth]{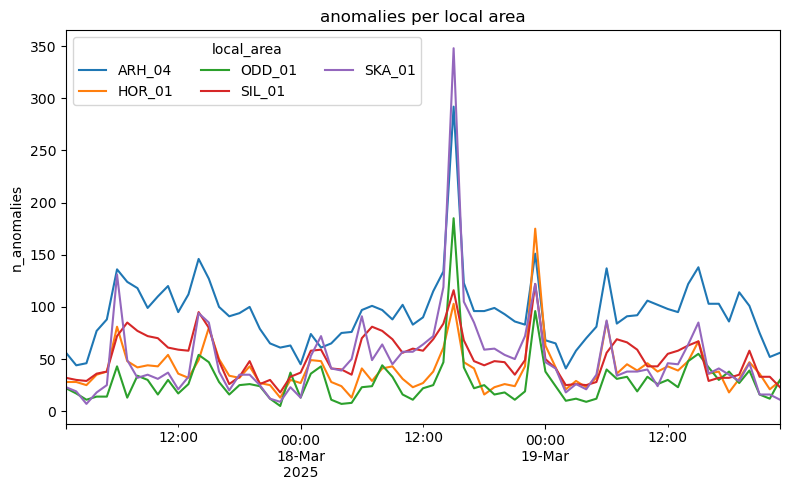}
   
    \caption{RAN (each line is a local\_area).}
     \label{subfig:ran_agg}
  \end{subfigure}

  \caption{\emph{Aggregated number of anomalies} in specific time windows on the EPC (top) and
  RAN (bottom) datasets. In both cases each curve corresponds to a network
  element (host in EPC, local\_area in RAN). Peaks indicate periods where
  multiple network elements are simultaneously anomalous and are used to
  prioritise operator attention before drilling down into specific counters
  and elements.}
  \label{fig:priority}
\end{figure}

Figure~\ref{fig:priority} shows example time series of these aggregated signals
on EPC (top) and RAN (bottom), grouped as indicated in the figure caption.
In both domains, short-lived spikes correspond to periods where many counters
are simultaneously flagged across one or more NEs and typically align with
operationally relevant events (e.g., configuration changes, signalling storms),
whereas a low, noisy baseline of occasional small peaks is treated as normal
background variability.

The resulting workflow is two-stage: (i) operators monitor the aggregated
priority stream to identify periods of interest, and (ii) for selected
peaks they drill down into NE- and counter-level views to investigate
root causes. This helped focus attention on the most severe or unusual
episodes, while allowing low-level background anomalies to be treated as
benign unless they clustered in time or across counters.

\subsubsection{\textbf{Illustrative Incidents}}

We now summarise several real incidents where the model’s alarms were
analysed by domain experts, with an overview provided in
Table~\ref{tab:incidents_ran_core}. In all cases, C-MTAD-GAT was run in the
unsupervised mode described earlier, with validation-based thresholds and
no access to ground-truth labels.

\begin{figure}[t]
\centering
\includegraphics[width=0.8\linewidth]{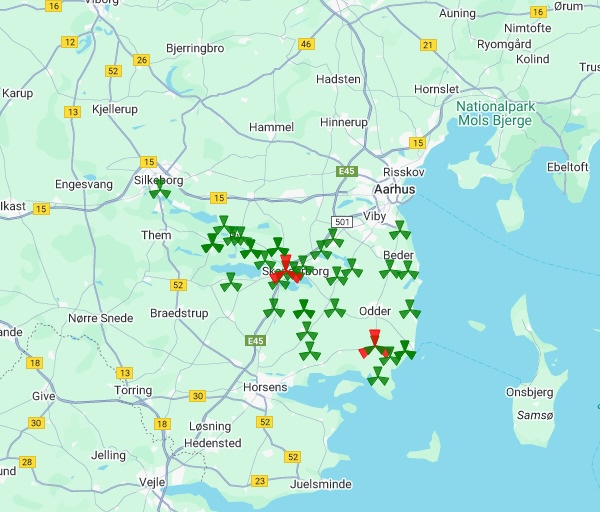}
\caption{Spatial footprint of the RAN incident in Fig.~\ref{subfig:ran_agg}:
sites with degraded stability ($\approx$~30 affected base stations in a single
RAN area) are highlighted on the map, illustrating how the aggregated
number of anomalies localises transmission-related problems to a specific
geographical cluster of network elements.}
\label{fig:pic_ran_story}
\end{figure}

\paragraph{Transmission-related instability in one RAN area}
The model highlighted a pronounced spike in the aggregated number of anomalies
localised to a specific RAN area (Fig.~\ref{subfig:ran_agg} and
Fig.~\ref{fig:pic_ran_story}). Elevated anomaly scores across approximately
30 sites within the same \texttt{local\_area} indicated widespread
instability. Investigation showed that a fault in the external transmission
provider’s network intermittently disrupted backhaul connectivity, degrading
stability across the affected sites. After the transmission fault was
resolved, KPIs and traffic levels returned to normal and the aggregated
anomaly number of anomalies dropped back to its baseline, with no further customer-visible
impact.

\begin{figure}[t]
\centering
\includegraphics[width=0.9\linewidth]{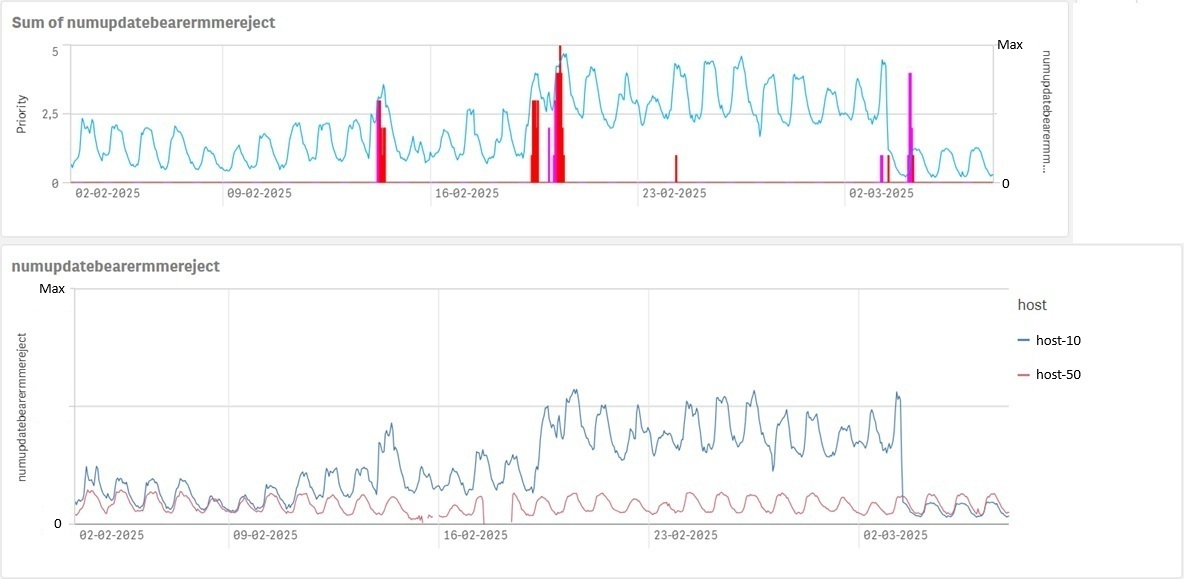}
\caption{Detected anomalies in the
\texttt{Update Bearer MME Reject}
counter. Top: aggregated anomaly priority score (red / purple bars) and sum of
the counter across hosts (line); purple bars mark anomalies within
maintenance windows. Bottom: per-host counter values over time.}
\label{fig:pic_story1}
\end{figure}

\paragraph{Update Bearer MME Reject increase (EPC)}
The model also flagged the onset of an extended period with elevated values
of the \texttt{Update Bearer MME Reject} counter
(Fig.~\ref{fig:pic_story1}). The aggregated anomaly priority score showed distinct
spikes aligned with increases in this counter across hosts. Domain experts
traced the issue to inbound roamers triggering failed modifications of
Quality of Service Class Identifier (QCI) and Allocation and Retention
Priority (ARP) values, leading to rejected bearer updates. The problem was
not yet visible through customer complaints, so the anomaly acted as an
early warning; after adjusting the MME configuration to enforce static
parameters for roamers, both the counter and the anomaly signal returned to
normal levels.

\begin{figure}[t]
\centering
\includegraphics[width=0.9\linewidth]{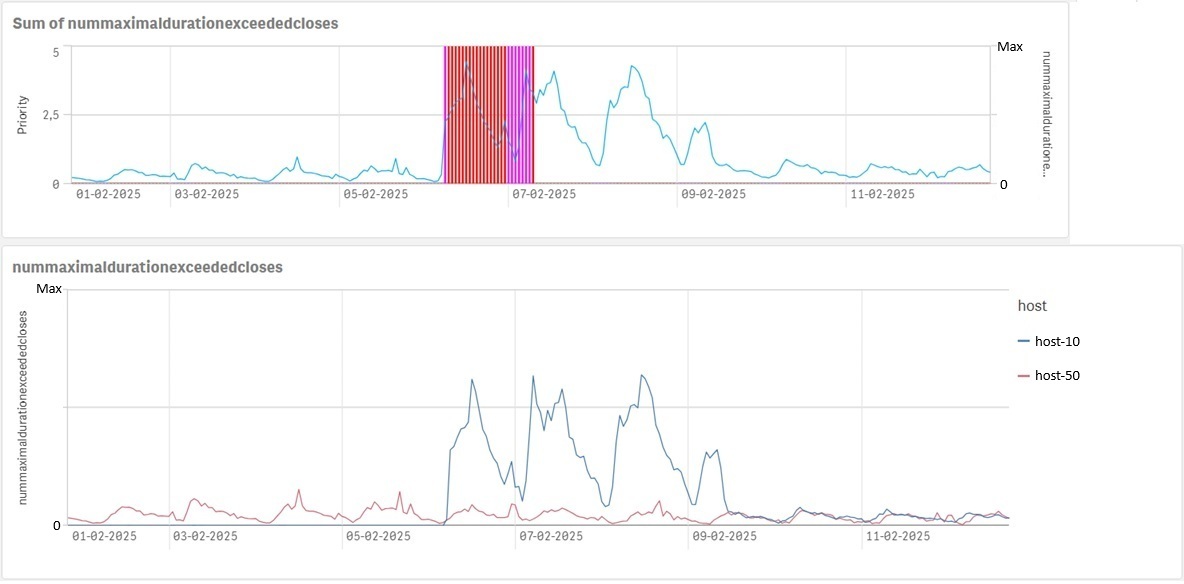}
\caption{Anomalies in the \texttt{Maximal Durations Exceeded} counter,
linked to unintended VoLTE call-duration limits. Top: aggregated anomaly
priority (bars); purple bars mark anomalies within
maintenance windows. and counter sum across hosts (line). Bottom: per-host
counter trajectories.}
\label{fig:pic_story2}
\end{figure}

\paragraph{VoLTE call-duration limitations (EPC)}
In another case, the model surfaced anomalies in the
\texttt{Maximal Durations Exceeded} counter
(Fig.~\ref{fig:pic_story2}), indicating an unexpected rise in sessions
terminated by reaching a maximal-duration limit. This was traced to a
misconfiguration that imposed stricter VoLTE call-duration limits on the
new MME platform than on the previous one. The anomaly thus acted as a
regression detector after a platform change; once the configuration was
corrected, the counter and associated anomalies decreased to their previous
baseline.

\begin{figure}[t]
\centering
\includegraphics[width=0.9\linewidth]{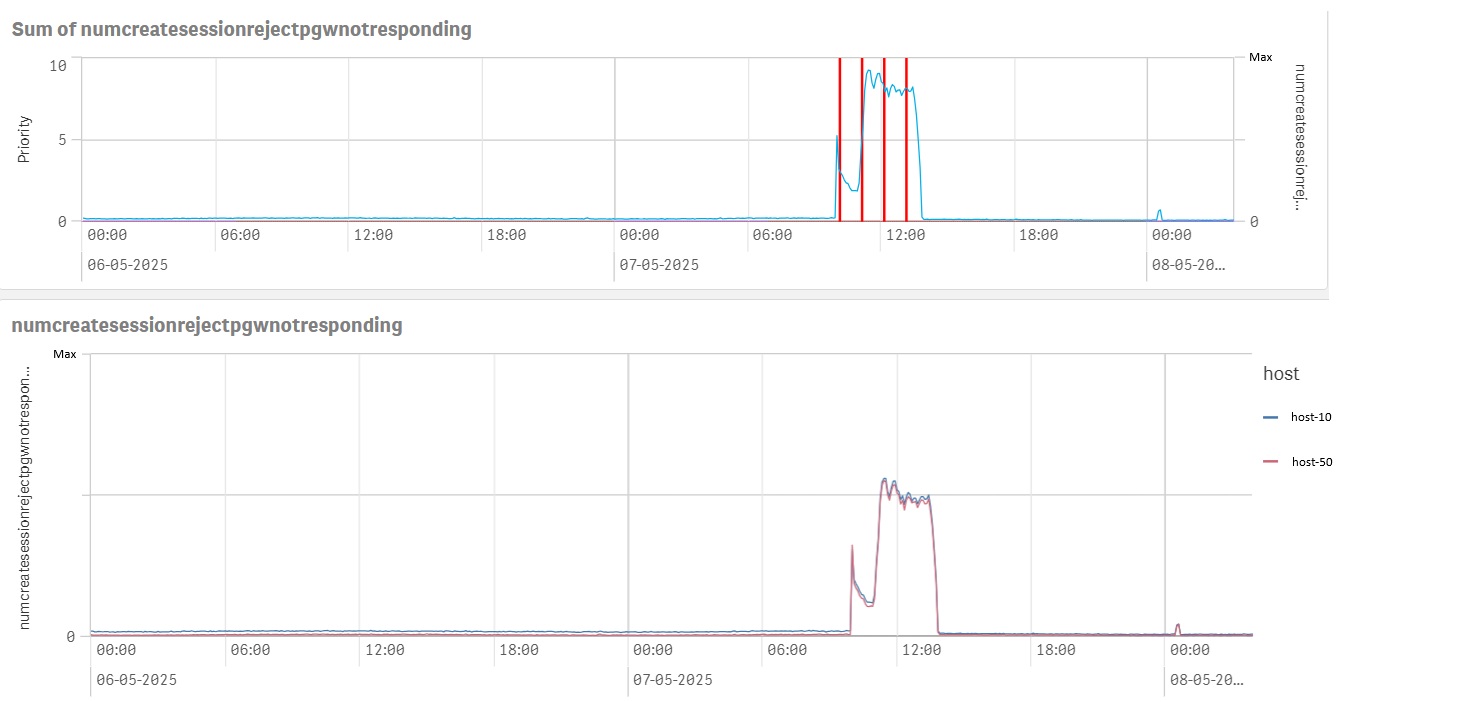}
\caption{Anomalies in the \texttt{Number of create session reject due to
PGW not responding} counter during a public-warning event. Top:
aggregated anomaly priority (bars) and counter sum across hosts (line).
Bottom: per-host counter values revealing the transient signalling
storm.}
\label{fig:pic_story4}
\end{figure}

\paragraph{Create Session Reject due to PGW not responding (EPC)}
The model also detected a sharp increase in the
\texttt{Number of create session reject due to PGW not responding}
counter (Fig.~\ref{fig:pic_story4}). This coincided with a mass
cell-broadcast / public-warning event that caused many users to
simultaneously wake their phones, creating a surge in idle-to-connected
transitions. The resulting spike in bearer-modification requests overloaded
parts of the EPC signalling path; response delays and timer expiries
triggered session release and re-creation, effectively creating a
``signalling storm''. Memory-allocation tuning, combined with the natural
decay of the broadcast effect, brought the counter back to normal, and the
anomaly view helped correlate the symptom with the temporal pattern of the
event.

\begin{figure}[t]
\centering
\includegraphics[width=0.9\linewidth]{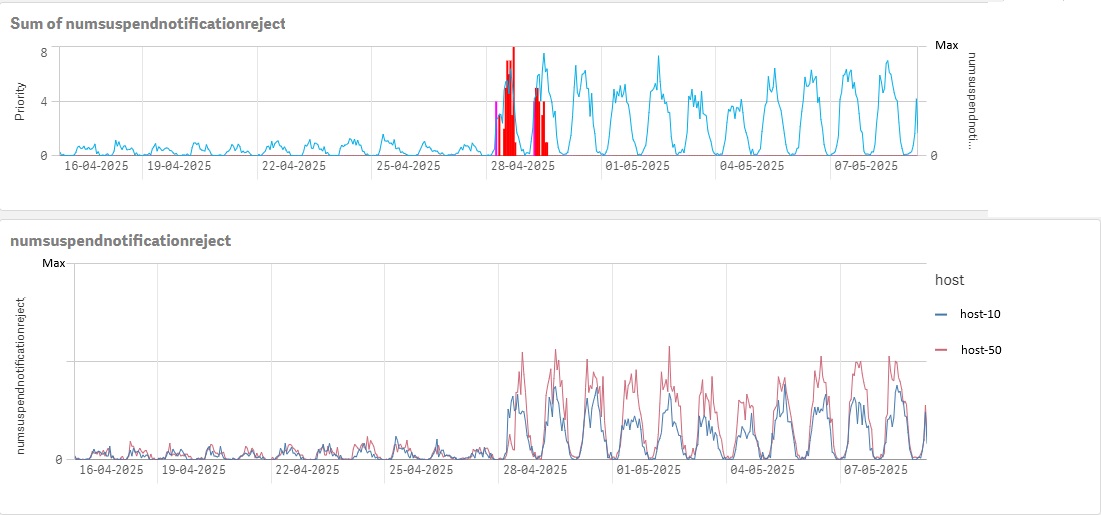}
\caption{Anomalies in the \texttt{Number of Suspend Notification Reject}
counter following a planned configuration change. Top: aggregated
anomaly priority (bars) and counter sum across hosts (line). Bottom:
per-host counter trajectories; purple bars indicate anomalies within
maintenance windows, confirming a deliberate shift in the operational
baseline rather than a fault.}
\label{fig:pic_story5}
\end{figure}

\paragraph{Suspend Notification Reject after configuration change (EPC)}
A further anomaly cluster involved an increase in the
\texttt{Number of Suspend Notification Reject} counter
(Fig.~\ref{fig:pic_story5}). Domain experts linked this to a planned
configuration change intended to reduce signalling load caused by a virtual
service provider’s Remote Authentication Dial-In User Service (RADIUS)
server (an MVNO integration). The timing of the anomalies matched the
maintenance window (purple bars), indicating that the model was reacting to
a new steady-state behaviour rather than an unexpected fault. The new
configuration proactively rejected specific requests when the provider did
not respond promptly, shifting the expected baseline of this counter;
here the anomaly view was mainly useful as a marker of a changed operating
regime rather than a business-critical incident.

\subsubsection{\textbf{Operational lessons}}
Across these incidents, several patterns emerge.
First, fully unsupervised anomaly detection is \emph{operationally useful}
even without labels: the model consistently surfaced network behaviours
that engineers deemed relevant, either as early warnings, regression
detectors, or markers of new baselines.
Second, human-in-the-loop validation is essential to turn raw anomaly
streams into business value. Expert feedback is needed to distinguish
between mathematically significant deviations and genuinely
business-critical issues, and to re-interpret persistent anomalies as
new ``normal'' regimes once planned changes take effect.
Third, the case studies highlight the importance of monitoring baseline
shifts and deciding when to retrain. In a live network, configuration
changes and traffic evolution regularly move counters to new operating
ranges; anomalies should both draw attention to such shifts and then
inform subsequent model updates, for instance via scheduled retraining
or unsupervised error-drift monitoring.
Finally, engineers consistently asked for tighter integration of
model output with existing tooling, e.g., root-cause hints and
direct surfacing of high-priority anomalies in NOC ticketing systems.
These requirements inform the future work outlined in the conclusion.

\section{Conclusion}

We have presented C-MTAD-GAT, a context-aware extension of MTAD-GAT that
combines GATv2 attention, conditional convolutions over categorical
metadata, and a deterministic GRU-AE reconstruction head for unsupervised
anomaly detection in telecom time series. On the public TELCO benchmark,
C-MTAD-GAT consistently outperforms MTAD-GAT, a $\beta$-MTAD-GAT variant,
and the Telco-specific DC-VAE under a common unsupervised calibration
protocol, achieving higher event-level and timestamp-level F1 while
raising fewer alarms. On large-scale RAN and EPC datasets from a national
operator, ablations show that C-MTAD-GAT backbone is more robust than MTAD-GAT
variants, that injecting context is particularly beneficial on
heterogeneous RAN data, and that a single centralised model trained over
thousands of NEs offers a favourable trade-off between scalability and
per-NE performance. Our scalability and Jaccard analyses further indicate
that the drift induced by centralisation is of the same order as the
intrinsic seed-to-seed variability of the model, rather than a
catastrophic loss of signal.

The operational case studies demonstrate that fully unsupervised anomaly
detection is feasible and useful in live telecom networks, even in the
absence of labels. C-MTAD-GAT systematically surfaced network behaviours
that engineers judged relevant—early warnings, regression detections, and
markers of new baselines—while also illustrating that statistically
significant deviations are not automatically business-critical. In
practice, value arises from how well the model helps engineers understand
multi-metric drift, relate anomalies to changes and incidents, and reason
about evolving operating regimes. Human-in-the-loop validation therefore
remains central: expert feedback is needed both to interpret alerts and to
decide when persistent deviations should be reclassified as normal
behaviour, informing change-aware monitoring and retraining schedules.

Future work will focus on tightening the loop between model output and
operations. Concretely, we plan to (i) incorporate lightweight root-cause
suggestion mechanisms, for example by ranking counters and NEs that
contribute most to each anomaly; (ii) integrate anomaly alerts and
aggregated priority scores directly into NOC ticketing workflows, so that
alerts can be triaged and closed within existing processes; and
(iii) explore adaptive retraining and error-drift monitoring strategies
that automatically detect when the model should be updated in response to
persistent baseline shifts. 


\section*{Acknowledgments}
This research was supported by the Research Council of Norway through the Machine Learning for Irregular Time Series (ML4ITS, grant no.\ 312062) and by SFI NorwAI (Centre for Research-based Innovation, grant no.\ 309834). SFI NorwAI is co-financed by the Research Council of Norway and its partners.

\bibliographystyle{IEEEtran}  
\bibliography{references}

\end{document}